\documentclass[journal]{IEEEtranTIE}
\usepackage{graphicx}
\usepackage{cite}
\usepackage{picinpar}
\usepackage{amsmath}
\usepackage{url}
\usepackage{flushend}
\usepackage[latin1]{inputenc}
\usepackage{colortbl}
\usepackage{soul}
\usepackage{multirow}
\usepackage{pifont}
\usepackage{color}
\usepackage{alltt}
\usepackage{enumerate}
\usepackage{siunitx}
\usepackage{breakurl}
\usepackage{epstopdf}
\usepackage{pbox}
\usepackage{amssymb}
\usepackage{multirow}
\usepackage{booktabs}   
\usepackage[caption=false,font=footnotesize,labelfont=rm,textfont=rm]{subfig}
\usepackage{float} 
\usepackage{amsthm}
\newtheorem{theorem}{Theorem}

\def\Higher#1{\textcolor[rgb]{0.0,0.0,0.0}{#1}}

\usepackage[pagebackref=false,breaklinks=true,colorlinks=true,bookmarks=false,citecolor=blue,linkcolor=black]{hyperref}

\begin{document}
\title{EDEN: Efficient Dual-Layer Exploration Planning for Fast UAV Autonomous Exploration in Large 3-D Environments}

\author{
	\vskip 1em
	Qianli Dong, Xuebo Zhang, Shiyong Zhang, Ziyu Wang, Zhe Ma, and Haobo Xi

\thanks{This work was supported in part by Natural Science Foundation of China under Grant 62303249, in part by the China Postdoctoral Science Foundation-Tianjin Joint Support Program under Grant 2023T013TJ, in part by the China Postdoctoral Science Foundation under Grant 2024M751526. (\textit{Corresponding author: Xuebo Zhang}.)

Qianli Dong, Xuebo Zhang, Shiyong Zhang, Ziyu Wang, Zhe Ma, and Haobo Xi are with the Institute of Robotics and Automatic Information System, College of Artificial Intelligence, Nankai University, Tianjin 300071, China, and also with the Engineering Research Center of Trusted Behavior Intelligence, Ministry of Education, and the Tianjin Key Laboratory of Intelligent Robotics, Nankai University, China (e-mail: qianlidong@mail.nankai.edu.cn; zhangxuebo@nankai.edu.cn; zhangshiyong@nankai.edu.cn; wziyu@mail.nankai.edu.cn; ma\_zhe@mail.nankai.edu.cn; xihaobo@mail.nankai.edu.cn)}
}
\maketitle
	
\begin{abstract}
Efficient autonomous exploration in large-scale environments remains challenging due to high planning computational cost and low-speed maneuvers. In this paper, we propose an Efficient Dual-layer Exploration plaNning method (EDEN). The insight of our dual-layer planning method is efficiently finding an acceptable long-term region routing and greedily exploring the target in the first region of routing at high speed. Specifically, we propose a long-term region routing approximate algorithm, called ``\Higher{exploration-oriented heuristic double-tree (EOHDT)} algorithm'', to ensure real-time planning in large-scale environments. Then, the viewpoint in the first routing region with the highest curvature-penalized score, which can effectively reduce decelerations caused by sharp turn motions, will be chosen as the next exploration target. To further speed up the exploration, we propose an aggressive and safe exploration-oriented trajectory planning approach to enhance exploration continuity and speed. The proposed method is compared with state-of-the-art methods in challenging simulation environments. The results show that the proposed method outperforms other methods in terms of exploration efficiency, computational cost, and trajectory speed. We also conduct real-world experiments to validate the effectiveness of the proposed method. The code will be open-sourced\footnote{\url{https://github.com/NKU-MobFly-Robotics/EDEN}}.
\end{abstract}


\markboth{IEEE TRANSACTIONS ON INDUSTRIAL ELECTRONICS}%
{}

\definecolor{limegreen}{rgb}{0.2, 0.8, 0.2}
\definecolor{forestgreen}{rgb}{0.13, 0.55, 0.13}
\definecolor{greenhtml}{rgb}{0.0, 0.5, 0.0}

\section{INTRODUCTION}
\IEEEPARstart{U}{nmanned} \Higher{aerial vehicles (UAVs) are popular in many 3-D exploration-relevant applications, e.g., search and rescue \cite{zhang2022fast, Shuang2024}, environment inspection \cite{Yao2024, Shuaizheng2023, Jie2023}, environment reconstruction \cite{Feng2024, Xuetao2024, Mingjie2024, xuetao2022}, healthcare services \cite{Bai2024, Bai2023}, among others.} However, most existing methods suffer from low efficiency, low-speed trajectories, and high computational cost.

Existing exploration methods can be coarsely separated into three kinds of methods: greedy methods, optimal-routing-based methods, and learning-based methods. Greedy methods attempt to select the target that maximizes the reward in a short period. The original frontier-based method \cite{Yamauchi1997} selects the closest frontier, i.e., the boundary voxel of explored and unknown space as the exploration target. To improve the UAV speed, the work in \cite{Cieslewski2017} proposes to select a frontier in the camera field-of-view (FOV) to minimize the velocity change. By combining next-best-views and rapid random tree (RRT), receding-horizon-next-best-views (RH-NBVP) \cite{Bircher2016} is proposed, and its variant methods \cite{Hongbiao2021, Zhong2022, Chaoqun2020, Dharmadhikari2020} show high exploration efficiency in small environments. However, these greedy methods suffer from massive back-and-forth maneuvers in large-scale environments due to the lack of long-term consideration. The optimal-routing-based methods \cite{Boyu2021, Yinghao2024, Boyu2023, Yichen2024, Song2018}, make decisions by constructing and solving routing problems to minimize the total routing length. The work in \cite{Song2018} proposes to find a coverage path by solving the traveling salesman problem (TSP). The method proposed by \cite{Boyu2021} constructs and solves the asymmetrical traveling salesman problem (ATSP) to find the shortest routing of frontiers.
To further reduce back-and-forth maneuvers, the approach in \cite{Yinghao2024} increases the routing priority of isolated frontiers. By solving routing problems, these methods dramatically reduce back-and-forth maneuvers. However, the computational cost grows fast with the number of routing nodes, which makes these methods have to execute stops to wait for planning results in large-scale environments. In addition, the works in \cite{Drew2024, Yichen2024} show that the speed of UAVs in exploration has great potential to be stimulated and can be significantly increased, while most existing state-of-the-art methods fail. \Higher{Some learning-based methods \cite{Cao2025, Cao2023, chaplotlearning} are proposed to explore the large and complex environments. However, they share the common limitation of suboptimal path planning due to discrete action spaces.} Therefore, a computation-efficient, far-sighted, and high-speed exploration method is highly expected.

Motivated by the above drawbacks of existing methods, we propose an $\textbf{E}$fficient $\textbf{D}$ual-layer $\textbf{E}$xploration pla$\textbf{N}$ning method (EDEN) which can plan efficient exploration motions in hundred-meter-level 3-D environments in real-time. Different from greedy methods and optimal-routing-based methods, the proposed dual-layer method first finds a long-term sub-optimal region routing efficiently and then decides the short-term targets that can fully take advantage of the current UAV speed. Specifically, we propose the \Higher{exploration-oriented heuristic double-tree (EOHDT)} algorithm to find a sub-optimal satisfactory routing for ATSP with low computational cost. Then, a viewpoint for safety and two viewpoints with the highest curvature-penalized exploration score in the first region of the routing will be selected for high-speed trajectory planning. Then, inspired by \cite{Tordesillas2022}, we propose an aggressive and safe exploration-oriented (ASEO) trajectory planning method for high-speed continuous exploration. The ASEO trajectory is composed of three parts: an exploring part, a continuous part, and a safety part. The time of the exploring part, the time of the continuous part, and the feasibility of the safety part are optimized simultaneously to compute an aggressive trajectory for fast exploration and a backup trajectory for safety.

We evaluate the proposed method by comparing it against state-of-the-art methods in various challenging environments. The results show that the proposed method has $5.2\%-89.7\%$ higher exploration efficiency, $10.4\%-69.7\%$ faster UAV speed, and $28.9\%-93.9\%$ lower computational cost. Further, the proposed method is deployed on a UAV and is tested in real-world environments to evaluate its effectiveness. The contributions of this work are summarized as follows:

\begin{enumerate}
\item 
We propose an \Higher{exploration-oriented heuristic double-tree (EOHDT)} algorithm for efficiently solving large-scale ATSP. \Higher{EOHDT} computes a satisfactory region routing with real-time performance even in hundred-meter-level environments.
\item To achieve high-speed continuous exploration, we propose a curvature-penalized continuous score evaluation for the determination of short-term exploration targets. 
\item We propose the ASEO trajectory planning method that ensures trajectory efficiency and safety simultaneously.
\item Extensive comparative simulations are conducted to evaluate the proposed method. We also test the proposed method by real-world experiments. We will release our source code.
\end{enumerate}

\section{RELATED WORK}
\subsection{Greedy Exploration Planning}
The frontier-based method is first proposed by Yamauchi \textit{et al.} \cite{Yamauchi1997}. It selects the frontier voxel that is closest to the robot as the exploration target. To minimize the UAV velocity change, the work in \cite{Cieslewski2017} selects the frontier voxel that is located at the boundary of the camera FOV as the exploration target. Different from the greedy frontier-based methods that only consider the next step exploration, Bircher \textit{et al}. \cite{Bircher2016} proposes RH-NBVP, which samples a set of exploration motion sequences and executes the most informative sequence's first step, and it shows high efficiency in short-term exploration. Nevertheless, in large-scale environments, it suffers from sample failure. To cope with this problem, Selin et al. \cite{Selin2019} use frontiers to guide sampling when a local sample fails. However, long-distance sampling is time-consuming, and it does not guarantee success even with the guidance of the frontier. To reduce the planning cost and improve long-distance exploration, road graphs, which can compress the complex explorable space information and traversable space information into compact nodes and edges, are widely used in recent proposed greedy methods \cite{Hongbiao2021, Batinovic2022, Chaoqun2020, xuetao2022, fsmp, Zezhou2023, Zhang2024}. Consequently, greedy methods are able to plan efficiently on road graphs in large environments. Although these methods show high efficiency in short-term exploration and low computational cost, they suffer from back-and-forth maneuvers due to a lack of long-term consideration.

\subsection{Optimal-Routing-Based Exploration Planning}
Optimal-routing-based methods usually formulate the exploration target decision problems into routing problems. Optimal-tour-based methods can be roughly devided into frontier-ATSP-based methods \cite{Boyu2021, Tang2023, Yinghao2024} and coverage-ATSP-based methods \cite{Shuang2025, Yichen2024, Song2018, Feng2024, Boyu2023}. Frontier-ATSP-based methods formulate ATSP using frontiers and find the shortest tour between frontiers. Coverage-ATSP-based methods formulate ATSP with unexplored space and find the shortest coverage path. \Higher{The main drawback of the optimal-routing-based methods is that they have to spend an extremely high computational cost on solving ATSP in large-scale environments.} Moreover, recent works \cite{fsmp,ericson2025information} report that optimal-routing-based methods usually do not have an obvious advantage in minimizing the exploration path distance compared to greedy methods. This is because optimal-routing-based methods usually make implicit optimal assumptions. Frontier-ATSP-based methods assume new frontiers will not arise after exploring a frontier. Coverage-ATSP-based methods assume all the unexplored space is traversable for the robot. Obviously, the above assumptions do not hold in most cases. Therefore, spending too much effort on computing a high-quality ATSP solution is not cost-effective.

\subsection{\Higher{Learning-Based Exploration Planning}}
\Higher{Learning-based methods have been used to tackle autonomous exploration problems with the advancement of machine learning. The work in \cite{li2019deep} presents a modular deep-reinforcement-learning-based exploration framework that uses a convolutional Q-network to select goals from partial maps. The works in \cite{gervet2023navigating, chaplotlearning} employ modular reinforcement learning frameworks for efficient visual exploration in unknown environments and provide systematic evaluations across realistic settings. The works in \cite{Cao2023,Cao2025} propose to construct a graph representation of the environment as the input to a graph neural network. However, they rely on discrete action spaces sampled from grid maps, which limits the optimality of the final path. In addition, the loading capacities of UAVs are usually limited, making them hard to carry extra devices for parallel computing.}

\subsection{Exploration-Oriented Trajectory Planning}
The exploration trajectory plays a crucial role in exploration tasks. The early proposed exploration methods \cite{Bircher2016, Yamauchi1997} use stop-and-go motions. To utilize the good maneuverabilities of UAVs, Dharmadhikari \textit{et al.} \cite{Dharmadhikari2020} adopts motion primitives to speed up the exploration. To further improve the trajectory speed, Zhou \textit{et al.} \cite{Boyu2021} optimizes a minimum-time trajectory with optimizable terminal velocity and acceleration for high-speed exploration. Although the minimum-time trajectory is very efficient, its optimizable terminal velocity and acceleration do not ensure improving the exploration continuity when new frontiers are generated after exploring the old frontier. What is more, the optimizable terminal velocity and acceleration can lead to a collision without considering obstacles that may be detected in the unknown space.

In summary, a far-sighted and computationally efficient exploration planning method and a trajectory planning method that enable fast continuous exploration while ensuring safety are required for exploration in large-scale environments.

\begin{figure}[!t]\centering
	\includegraphics[width=8.5cm]{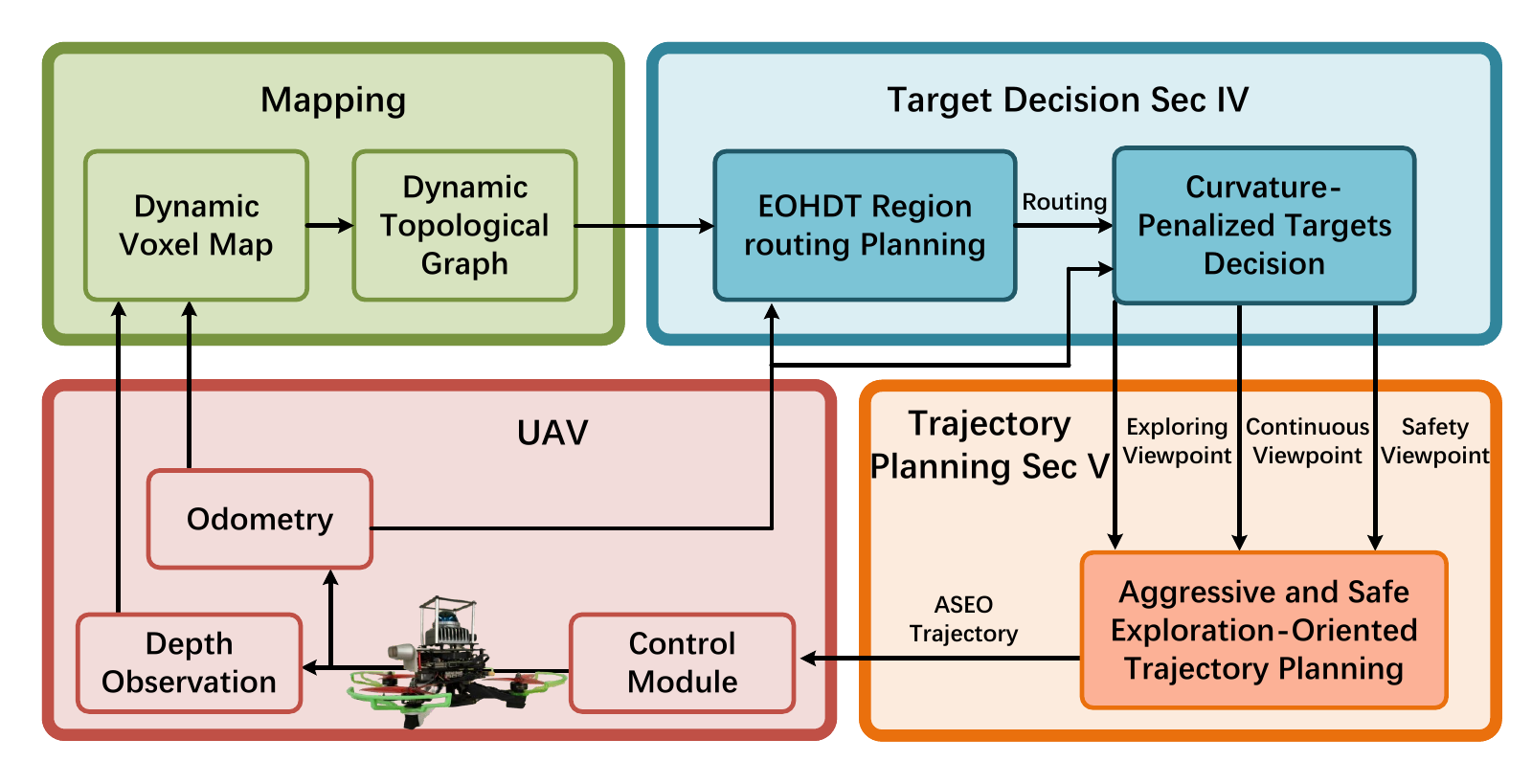}
	\caption{The overview of the proposed method framework.}
\label{sys_overview}
\end{figure}

\section{SYSTEM OVERVIEW}
\begin{figure}[!t]\centering
	\includegraphics[width=8.5cm]{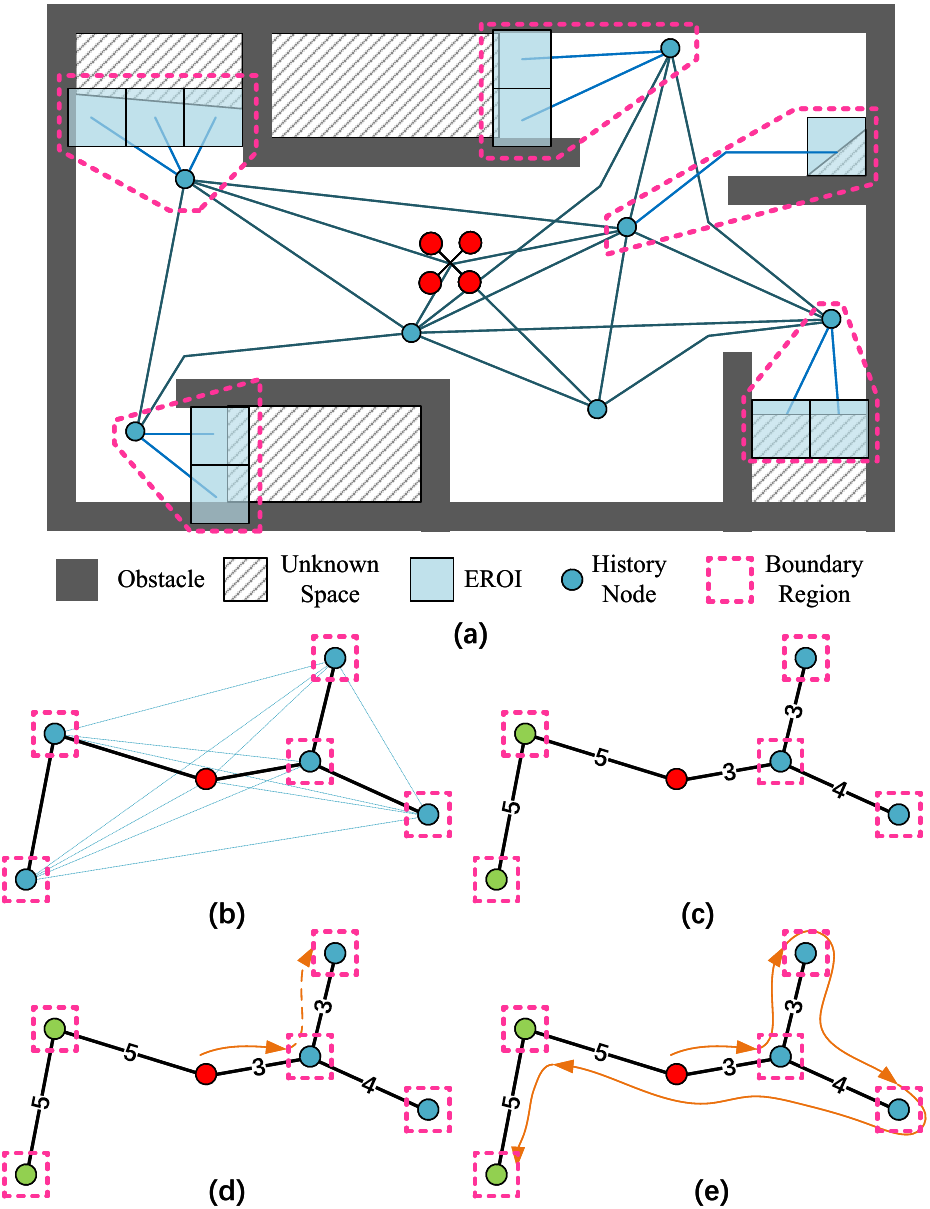}
	\caption{An illustration of DTG and an instance of \Higher{EOHDT}-based routing planning. (a) DTG and its boundary regions (pink dashed boxes). (b) An MST that connects all the \Higher{boundary regions} of DTG is extracted. (c) The leaf point, whose distance is the longest to the root node, and its corresponding branch (green) will be executed last. The numbers indicate the lengths of edges. (d) Once there are multiple child nodes for the current node, the child node (orange dashed curve) whose first node has the shortest distance to the current node will be chosen as the next branch. (e) The final routing found by the proposed \Higher{EOHDT}-based routing planning.}
\label{routing}
\end{figure}

The dual-layer exploration target decision module plans both the long-term region routing and the short-term viewpoint targets. The trajectory planning module is responsible for planning high-speed continuous exploration trajectories. To achieve fast decision-making, a dynamic voxel map with resolution $r_v$ is updated by integrating depth image and odometry information. Different from the normal voxel map, the dynamic voxel map only maintains a local part of voxels around the UAV, while other voxels are dynamically read and written to save memory usage in large environments. Based on the voxel map, a dynamic topological graph (DTG, Fig. \ref{routing}(a)) is maintained to manage exploration-related information. The traversable information is maintained by DTG's history nodes and edges between them, and the explorable space information is maintained by DTG's explorable region of interest (EROI), which is a set of uniform-scale cubic regions that can be explored by visiting corresponding active viewpoints. For detailed information, we refer readers to our previous work \cite{dong2024fast}. The exploration planning utilizes DTG to efficiently formulate ATSP, and ATSP is solved through \Higher{EOHDT} to find satisfactory exploration region routing with small computation. Then, three viewpoints, i.e., an exploring viewpoint, a continuous viewpoint, and a safety viewpoint in the first region of the region routing, are selected for the following trajectory planning. With these three viewpoints, an aggressive and safe exploration-oriented trajectory is generated, which is divided into three parts from the exploring viewpoint. The first part is named the exploring part, which goes from the UAV's current position to the exploring viewpoint. The second part is named the continuous part, which goes from the exploring viewpoint to the continuous viewpoint. The last part is named the safety part, which goes from the exploring viewpoint to the safety viewpoint. After every replanning, the exploring part of the aggressive and safe exploration-oriented trajectory will be actually executed.

\section{DUAL-LAYER EXPLORATION VIEWPOINTS PLANNING}
In this section, we first propose the \Higher{EOHDT} algorithm for finding a satisfactory region routing efficiently in large-scale environments. Then, we use a curvature-penalized continuous score to decide the targets in the first region of the tour for the following high-speed ASEO trajectory planning.

\subsection{EOHDT-Based Global Routing Planning}
\Higher{As analyzed in Section II, a far-sighted and computationally efficient exploration planner for the large-scale exploration tasks is required.
Therefore, instead of finding a high-quality routing of all precise explorable frontiers, the proposed method aims to spend low computation on finding a satisfactory coarse routing of boundary regions. We propose the EOHDT-based routing planning, which has a low computational complexity and a length guarantee in the worst case, for solving ATSP.} The process of \Higher{EOHDT}-based routing planning is shown in Fig. \ref{routing}(b-e).

\Higher{Firstly, we extract all the boundary regions. Here, a boundary region is composed of a history node and a set of EROIs (at least one) connected to this history node as shown in Fig. \ref{routing}(a).} To reduce the computation when formulating ATSP, the distances between all \Higher{boundary regions} and the distances between the robot and all \Higher{boundary regions} are computed by conducting the Dijkstra algorithm on DTG. Thanks to the sparse DTG, the ATSP can be formulated efficiently.

To solve ATSP efficiently, we design an \Higher{EOHDT} algorithm, which is improved from the double-tree algorithm \cite{Christofides} and tailored to ATSP. The key idea of the double-tree algorithm is to construct a minimum spanning tree (MST, Fig. \ref{routing}(b)) using the Prim algorithm and find a walk that goes around the MST. The solution routing of the double-tree algorithm ensures no longer than twice the optimal routing. However, the double-tree algorithm is designed for symmetric TSP. For exploration-oriented ATSP, the robot does not have to return to the original node as stated by \cite{Boyu2021}. Therefore, to minimize the routing length, the leaf node that is the farthest from the robot and nodes on its corresponding branch (in Fig. \ref{routing}(c), the green nodes) will be executed last to save the highest cost of returning to the original node. Another problem is how to decide the sequence of the other branches. Since all the regions will be explored finally, the proposed method chooses the next node closest to the current node heuristically when the current node has multiple child nodes, as shown in Fig. \ref{routing}(d). Finally, a routing for ATSP will be found as shown in Fig. \ref{routing}(e).
\begin{theorem}
\Higher{The EOHDT runs in $\mathcal{O}(n^2)$ time, where $n$ denotes the number of the history nodes.}
\end{theorem}
\begin{proof}
\Higher{EOHDT starts by generating an MST utilizing the Prim algorithm, which takes $\mathcal{O}(n^2)$ time. Then, EOHDT finds the farthest leaf node among $n_{leaf}$ leaf nodes, which takes $\mathcal{O}(n_{leaf})$ time. Let $m$ be the number of nodes that have multiple child nodes in the MST. Then, EOHDT needs to conduct $m$ times sorting of child nodes. It results in $\mathcal{O}(\sum_{i=1}^{m}n_i\log n_i)$ time complexity, where $n_i$ is the child nodes number of the $i$-th node among the $m$ nodes. Because $n_i < n$ and $2\leq n_i$, we have $\sum_{i=1}^{m} n_i\log n_i < n^2$.}
\Higher{Therefore, EOHDT runs in $\mathcal{O}(n^2+n_{leaf}+\sum_{i=1}^{m}  n_i\log n_i)\approx\mathcal{O}(n^2)$ time.}
\end{proof}
\begin{theorem}
\Higher{The length of the routing solved by EOHDT $l_{E}$ holds $l_{E}\leq 2l^*-l_m$, where $l^*$ is the length of the optimal solution of the ATSP, and $l_m$ is the longest distance among\ distances between the robot and all the nodes on the MST.}
\end{theorem}
\begin{proof}
\Higher{Given the summation of all the edge lengths of the MST $l_{MST}$, it holds $l_{MST}\leq l^*$. It is easy to derive}
\begin{align}
\label{2mst}
\Higher{l_{E} = 2l_{MST}-l_m\leq 2l^*-l_m.}
\end{align}
\end{proof}

\subsection{Curvature-Penalized Viewpoints Planning}
After finding the long-term routing of regions, three viewpoints (an exploring viewpoint, a continuous viewpoint, and a safety viewpoint) in the first region of the routing will be selected for ASEO trajectory planning. The UAV will go to explore the exploring viewpoint (in the explored map) and maintain a speed and an acceleration conducive to exploring the continuous viewpoint (in the explored map or unexplored map) if no obstacle is detected, otherwise it can return to the safety viewpoint (in the explored map).

For detailed viewpoint evaluation, existing methods usually use the path distance and turning angle to avoid long traveling distances and sharp turns in a decoupled manner. However, the maximum velocity is limited by the curvature of the trajectory at a point according to the circular motion model. The maximum velocity $v_s(\mathbf{p})$ at $\mathbf{p}$ is computed by
\begin{align}
\label{v-curve}
v_{s}(\mathbf{p}) = \min\left(\sqrt{\frac{\lambda_a\cdot a_m} {K_c(\textbf{p})}}, v_m\right),
\end{align}
where $v_m$ is the maximum linear velocity of the UAV, $a_m$ is the maximum linear acceleration of the UAV, $K_c(\mathbf{p})$ is the curvature of the trajectory at $\mathbf{p}$, \Higher{and $\lambda_a \in (0.0,1.0)$ is a tunable parameter for reducing sharp turns with high accelerations. In the following viewpoints planning, a smaller $\lambda_a$ leads to fewer turns but may miss the opportunities to explore some corner areas.} Based on this fact, we designed a curvature-penalized motion score evaluation method to select the best exploring target and the best continuous target for high-speed exploration.

\subsubsection{\bf{Curvature-Penalized Exploring Viewpoint and Continuous Viewpoint Planning}} 
An exploring viewpoint in the first region of routing and a continuous viewpoint located in the corresponding EROI of the exploring viewpoint will be selected for aggressive high-speed trajectory planning. 

First, a Dijkstra search, which starts from the current UAV position in the explored space, is conducted to find the path to each candidate viewpoint in the first region of the routing. \Higher{Then, for each path to a candidate viewpoint, an anchor point on the path, which is $1.0\,s\times v_m$ length away from the UAV or the path terminal if the path is shorter than $1.0\,s\times v_m$, is selected to formulate a virtual circular arc trajectory.} \Higher{The formulation of a virtual circular arc trajectory to the $j$-th candidate viewpoint $\mathbf{vp}_j=\left\{\mathbf{p}_j, \psi_j\right\} \in \mathbb{R}^3\times [-\pi,\pi)$ is constrained by the current UAV position $\mathbf{p}_u \in \mathbb{R}^3$, the current UAV speed $\dot{\mathbf{p}}_u \in \mathbb{R}^3$, and the anchor point $\mathbf{p}_{a,j} \in \mathbb{R}^3$. The curvature $K_{u,j}$ of the virtual circular arc trajectory can be computed by}
\begin{align}
\nonumber
\theta = \arccos&(\frac{\dot{\mathbf{p}}_u^\mathrm{T}\cdot(\mathbf{p}_{a,j} - \mathbf{p}_u)}{\Vert\dot{\mathbf{p}}_u\Vert_2\cdot\Vert\mathbf{p}_{a,j} - \mathbf{p}_u\Vert_2}),\\
K_{u,j}&=\Higher{\frac{2\sin(\theta)}{\Vert\mathbf{p}_{a,j} - \mathbf{p}_u\Vert_2}.}
\end{align}
The virtual velocity $v_{as,j}$ to $\mathbf{vp}_j$ is penalized by the curvature $K_{u,j}$ and is computed by Eq. \ref{v-curve}. The motion cost of going to this exploring target is computed by
\begin{align}
\mathrm{CostE}(\mathbf{vp}_j)=\max\left(\frac{\mathrm{PL}(\mathbf{p}_j, \mathbf{p}_u)}{v_{as,j}}, \frac{|\mathrm{YD}(\psi_j, \psi_u)|}{\dot{\psi}_m}\right),
\end{align}
\Higher{where the function $\mathrm{PL}(\mathbf{p}_j$, $\mathbf{p}_u)$ computes the length of the path from $\mathbf{p}_u$ to $\mathbf{p}_j$, $\mathrm{YD}(\psi_j, \psi_u):\left[-\pi,\pi\right)\times[-\pi,\pi)\mapsto[-\pi,\pi)$ computes the yaw difference between the current UAV yaw $\psi_{u}$ and $\psi_{j}$ by angle unwrapping, and $\dot{\psi}_m$ is the maximum yaw velocity.}
Since the terminal velocity at $\mathbf{vp}_j$ is unknown, we continue to use the $v_{as,j}$ for computing the continuous motion cost from $\mathbf{vp}_j$ to the $k$-th viewpoint $\mathbf{vp}_k=\left\{\mathbf{p}_k, \psi_k\right\} \in \mathbb{R}^3\times [-\pi, \pi)$ inside the EROI of $\mathbf{vp}_j$
\begin{align}
\mathrm{CostC}(\mathbf{vp}_k)=\max\left(\frac{\mathrm{PL}(\mathbf{p}_k, \mathbf{p}_j)}{v_{as,j}}, \frac{|\mathrm{YD}(\psi_k, \psi_j)|}{\dot{\psi}_m}\right).
\end{align}
\Higher{Finally, the score for evaluating the viewpoint pair is computed by}
\begin{align}
\Higher{\mathrm{G}(\mathbf{vp}_k,\mathbf{vp}_j)=\mathrm{e}^{-\lambda_e\mathrm{CostE}(\mathbf{vp}_j)}\cdot(1+\mathrm{e}^{-\lambda_e\mathrm{CostC}(\mathbf{vp}_k)}).}
\end{align}
\Higher{Here, $\lambda_e$ is a positive tunable parameter. For a larger $\lambda_e$, the planner is inclined to choose the exploring viewpoint with a low $\mathrm{CostE}(\mathbf{vp}_j)$. For a smaller $\lambda_e$, the planner is inclined to choose the viewpoint pair with a small total motion cost.}
\Higher{The viewpoint-pair with the highest score is selected as the exploring viewpoint $\mathbf{vp}_e=\{\mathbf{p}_e,\psi_e\}$ and the continuous viewpoint $\mathbf{vp}_c=\{\mathbf{p}_c,\psi_c\}$.}

\Higher{It is worth noting that the voxels on the line between $\mathbf{p}_j$ and $\mathbf{p}_k$ can not be occupied but are free and unknown. This ensures that $\mathbf{p}_k$ can be observed at $\mathbf{p}_j$, which makes it possible to explore continuously.} To efficiently check the feasibility of the viewpoint pairs, we propose a continuous line-checking method that follows the fast collision checking fashion \cite{Hou_2025} to avoid recalculation of indices of voxels at overlapping positions of continuous lines. Specifically, our method generates a table of indices of voxels that need to be checked in advance and iterates through the table when checking an EROI. Once an iterative voxel is found occupied, its corresponding viewpoint pairs are confirmed infeasible and will not be considered.

\subsubsection{\bf{Safety Viewpoint Planning}} 
After finding $\mathbf{vp}_e$ and $\mathbf{vp}_c$, another viewpoint in the first region of routing will be selected. \Higher{The safety viewpoint $\mathbf{vp}_s=\{\mathbf{p}_s,\psi_s\}$ will be a feasible backup target, which ensures the UAV being able to return to the safe space, if new obstacles are detected on the trajectory between $\mathbf{vp}_e$ and $\mathbf{vp}_c$. Without $\mathbf{vp}_s$, the UAV could be too fast to brake at $\mathbf{vp}_e$. Although we could find the optimal safety viewpoint through solving another ATSP starting from $\mathbf{vp}_e$, we directly select the viewpoint closest to $\mathbf{vp}_e$ as $\mathbf{vp}_s$ to save the computation. This safety viewpoint is found by conducting a Dijkstra search starting from $\mathbf{vp}_e$ inside the free space. Once a viewpoint is searched, $\mathbf{vp}_s$ is set as this viewpoint, and the Dijkstra search stops.}

\begin{figure}[!t]\centering
	\includegraphics[width=7.8cm]{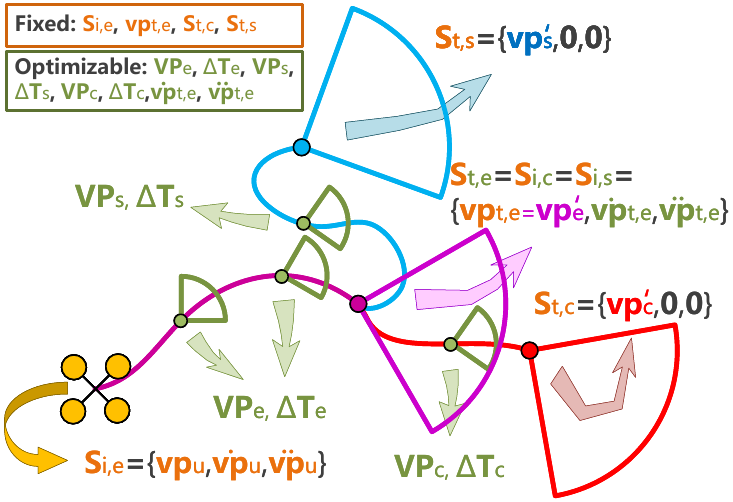}
	\caption{An instance of ASEO trajectory. The initial position, velocity, and acceleration of $\mathcal{J}_e$ (violet) are fixed with the current UAV position, velocity, and acceleration (yellow). The continuous trajectory $\mathcal{J}_c$ (red) ends at $\mathbf{vp}_c'$. \Higher{The safety trajectory $\mathcal{J}_s$ (blue) ends at $\mathbf{vp}_s'$.} The optimizable parameters, including intermediate 4-D points, trajectory durations, velocity, and acceleration at $\mathbf{vp}_e'$, are illustrated in green.}
\label{ASEO}
\end{figure}

\section{ASEO TRAJECTORY OPTIMIZATION}
Inspired by the fast and safe trajectory planner (FASTER) \cite{Tordesillas2022}, we propose to optimize our 4-D aggressive and safe exploration-oriented (ASEO) trajectory, which is based on the MINCO trajectory \cite{Zhepei2022}. Different from the 3-D MINCO trajectory, the fourth dimension is the yaw, and we optimize it with the position trajectory together.

A 4-D MINCO trajectory \Higher{$\mathcal{J}(t):\left[t_s,t_e\right]\mapsto\mathbb{R}^4$} ($t_s$ and $t_e$ are the start time and end time of the trajectory) is paramerized by the initial state $\textbf{S}_i=\left\{\mathbf{vp}_i,\dot{\mathbf{vp}}_i,\ddot{\mathbf{vp}}_i\right\}$, the terminal state $\mathbf{S}_t=\left\{\mathbf{vp}_t,\dot{\mathbf{vp}}_t,\ddot{\mathbf{vp}}_t\right\}$, the $m-1$ intermediate viewpoints $\mathbf{VP}=\left\{\mathbf{vp}_1,\mathbf{vp}_2,\cdots,\mathbf{vp}_{m-1}\right\}$, and $m$ durations of trajectory segments $\Delta\mathbf{T}=\left\{\Delta t_{1},\Delta t_{2},\cdots,\Delta t_{m}\right\}$. As shown by Fig. \ref{ASEO}, the proposed ASEO trajectory is composed of three connected 4-D MINCO trajectories: an exploring trajectory $\mathcal{J}_e$ (paramerized by $\mathbf{S}_{i,e}$, $\mathbf{S}_{t,e}$, $\mathbf{VP}_e$, and $\Delta\mathbf{T}_e$), a continuous trajectory $\mathcal{J}_c$ (paramerized by $\mathbf{S}_{i,c}$, $\mathbf{S}_{t,c}$, $\mathbf{VP}_c$, and $\Delta\mathbf{T}_c$), and a safety trajectory $\mathcal{J}_s$ (paramerized by $\mathbf{S}_{i,s}$, $\mathbf{S}_{t,s}$, $\mathbf{VP}_s$, and $\Delta\mathbf{T}_s$). \Higher{The three trajectories of ASEO are connected at the exploring viewpoint $\mathbf{vp}_e$ to ensure the coverage of unknown space, which is different from FASTER \cite{Tordesillas2022}, whose connecting point of its ``Whole'' trajectory and ``Safety'' trajectory can be any feasible point on the ``Whole'' trajectory.} Some parameters of the ASEO trajectory are fixed. \Higher{The initial state of the exploring trajectory $\mathbf{S}_{i,e}=\left\{\mathbf{vp}_{i,e},\dot{\mathbf{vp}}_{i,e},\ddot{\mathbf{vp}}_{i,e}\right\}$ is aligned with the UAV position $\{\mathbf{p}_c,\psi_c\}$, velocity $\{\dot{\mathbf{p}}_c,\dot{\psi}_c\}$, and acceleration $\{\ddot{\mathbf{p}}_c,\ddot{\psi}_c\}$. $\mathcal{J}_e$'s terminal pose $\mathbf{vp}_{t,e}$, $\mathcal{J}_c$'s initial pose $\mathbf{vp}_{i,c}$, and $\mathcal{J}_s$'s initial pose $\mathbf{vp}_{i,s}$ are fixed to the exploring viewpoint. $\mathcal{J}_c$'s terminal pose $\mathbf{vp}_{t,c}$ and $\mathcal{J}_s$'s terminal pose $\mathbf{vp}_{t,s}$ are fixed to the continuous viewpoint and the safety viewpoint correspondingly, their terminal velocity and acceleration are set to zero.} We summarize these fixed parameters as:
\begin{align}
\label{fixed}
\nonumber
&\mathbf{vp}_{i,e},\dot{\mathbf{vp}}_{i,e},\ddot{\mathbf{vp}}_{i,e}=\{\mathbf{p}_{c},\psi_{c}\},\{\dot{\mathbf{p}}_{c},\dot{\psi}_{c}\},\{\ddot{\mathbf{p}}_{c},\ddot{\psi}_{c}\},\\
\nonumber
&\mathbf{vp}_{t,e}=\mathbf{vp}_{i,c}=\mathbf{vp}_{i,s}=\mathbf{vp}_{e}'=\{\mathbf{p}_e,\psi_e'\},\\
\nonumber
&\mathbf{vp}_{t,c}=\mathbf{vp}_{c}'=\{\mathbf{p}_c,\psi_c'\}, \mathbf{vp}_{t,s}=\mathbf{vp}_{s}'=\{\mathbf{p}_s,\psi_s'\},\\
&\dot{\mathbf{vp}}_{t,c}=\ddot{\mathbf{vp}}_{t,c}=\dot{\mathbf{vp}}_{t,s}=\ddot{\mathbf{vp}}_{t,s}=\{\mathbf{0},0\}.
\end{align}
The rest of the parameters are optimizable, and these parameters are shown in Fig. \ref{ASEO} (green). To ensure smoothness, three trajectories share the same velocity and acceleration at the connected point. \Higher{It is worth noting that we use $\psi_e'$, $\psi_c'$, $\psi_s'\in\mathbb{R}$ instead of $\psi_e$, $\psi_c$, $\psi_s\in[-\pi,\pi)$ to avoid the rotations greater than $\pi$. Specifically, $\psi_e'$, $\psi_c'$, and $\psi_s'$ are calculated by $\psi_e'=\mathrm{YD}(\psi_e,\psi_u)+\psi_u$, $\psi_c'=\mathrm{YD}(\psi_c,\psi_e)+\psi_e'$, and $\psi_s'=\mathrm{YD}(\psi_s,\psi_e)+\psi_e'$. The intermediate yaws of $\mathcal{J}_e$ are bounded within $\left[\min(\psi_e', \psi_u)-\pi,\max(\psi_e', \psi_u)+\pi\right]$, the intermediate yaws of $\mathcal{J}_c$ are bounded within $\left[\min(\psi_c', \psi_e')-\pi,\max(\psi_c', \psi_e')+\pi\right]$, and the intermediate yaws of $\mathcal{J}_s$ are bounded within $\left[\min(\psi_s', \psi_e')-\pi,\max(\psi_s', \psi_e')+\pi\right]$.}

\Higher{Unlike FASTER, which optimizes different trajectories in a decoupled manner, we optimize the three trajectories of the ASEO trajectory simultaneously to ensure that each trajectory is feasible.} The objective function of the ASEO trajectory optimization problem is formulated as follows:
\begin{align}
\label{obj}
\textrm{J}_{ASEO}(\mathcal{J}_e,\mathcal{J}_c,\mathcal{J}_s)=j_{e,f}+j_{e,t}+j_{c,f}+j_{c,t}+j_{s,f},
\end{align}
where $j_{e,f},j_{c,f}$, and $j_{s,f}$ are the feasibility costs of $\mathcal{J}_e,\mathcal{J}_c$, and $\mathcal{J}_s$ correspondingly. To speed up the continuous exploration, $j_{e,t}$ and $j_{c,t}$ (the time costs of $\mathcal{J}_e$ and $\mathcal{J}_c$) are optimized. \Higher{The feasibility cost $j_{\mu,f}, \mu\in\{e,c,s\}$ is computed by
\begin{align}
\label{feasibility}
\nonumber
j_{\mu,f}&=\int_{t_s,\mu}^{t_{e,\mu}} \lambda_p\mathrm{Vp}_\mu(t)+\lambda_d(\mathrm{Vv}_\mu(t)+\mathrm{Va}_\mu(t)\\
&+\mathrm{V\dot{\psi}}_\mu(t)+\mathrm{V\ddot{\psi}}_\mu(t)) dt,
\end{align}
where $\mathrm{Vp}_\mu(t)$ calculates the position violation of collision-free corridor constraint, and $\mathrm{Vv}_\mu(t)$, $\mathrm{Va}_\mu(t)$, $\mathrm{V\dot{\psi}}_\mu(t)$, and $\mathrm{V\ddot{\psi}}_\mu(t)$ calculate the violations of linear velocity, linear acceleration, yaw velocity, and yaw acceleration, respectively.} \Higher{The penalty weights are $\lambda_p$ and $\lambda_d$. For the time cost $j_{\mu,t}, \mu\in\{e,c\}$, it is computed by}
\begin{align}
\label{time}
\Higher{j_{\mu,t}=\lambda_t\sum\limits_{i=1}^{m_\mu}\Delta t_{\mu,i},}
\end{align}
\Higher{where $\Delta t_{\mu,i} \in \Delta \mathbf{T}_\mu$, $m_\mu$ is the number of trajectory segments, and $\lambda_t$ is the penalty weight of time.} For details of derivation and optimization, we refer the readers to \cite{Zhepei2022}.

After generating the ASEO trajectory, $\mathcal{J}_e$ is executed. \Higher{Our exploration planner replans (replans viewpoints and regenerates ASEO trajectory) when it encounters the following conditions: (1) the distance from an obstacle to a point on $\mathcal{J}_e$, $\mathcal{J}_c$, or $\mathcal{J}_s$ is shorter than the radius of the UAV $r_u$; (2) $\mathbf{vp}_e$ does not cover any unknown space or the UAV is about to reach $\mathbf{vp}_e$ ($\Vert\mathbf{p}_u-\mathbf{p}_e\Vert_2<d_{th}$ and $|\mathrm{YD}(\psi_e,\psi_u)|<\psi_{th}$, where $d_{th}$ is the distance threshold and $\psi_{th}$ is the yaw threshold); (3) the UAV has executed $\mathcal{J}_e$ for a replanning period $t_r$ or $\mathcal{J}_e$ will expire within a replanning offset time $t_o$.}



\section{EXPERIMENTS and RESULTS}
\subsection{Implementation Details}
To evaluate the proposed method, we compare our method against state-of-the-art methods \cite{Yichen2024, Boyu2021} in simulation experiments and conduct real-world experiments.

\begin{figure}[htp]
\centering
\subfloat[UAV Platform.]{
\label{raw_vox_cells}
\begin{minipage}[htp]{0.32\linewidth}
\centering
\includegraphics[width=1.0\textwidth]{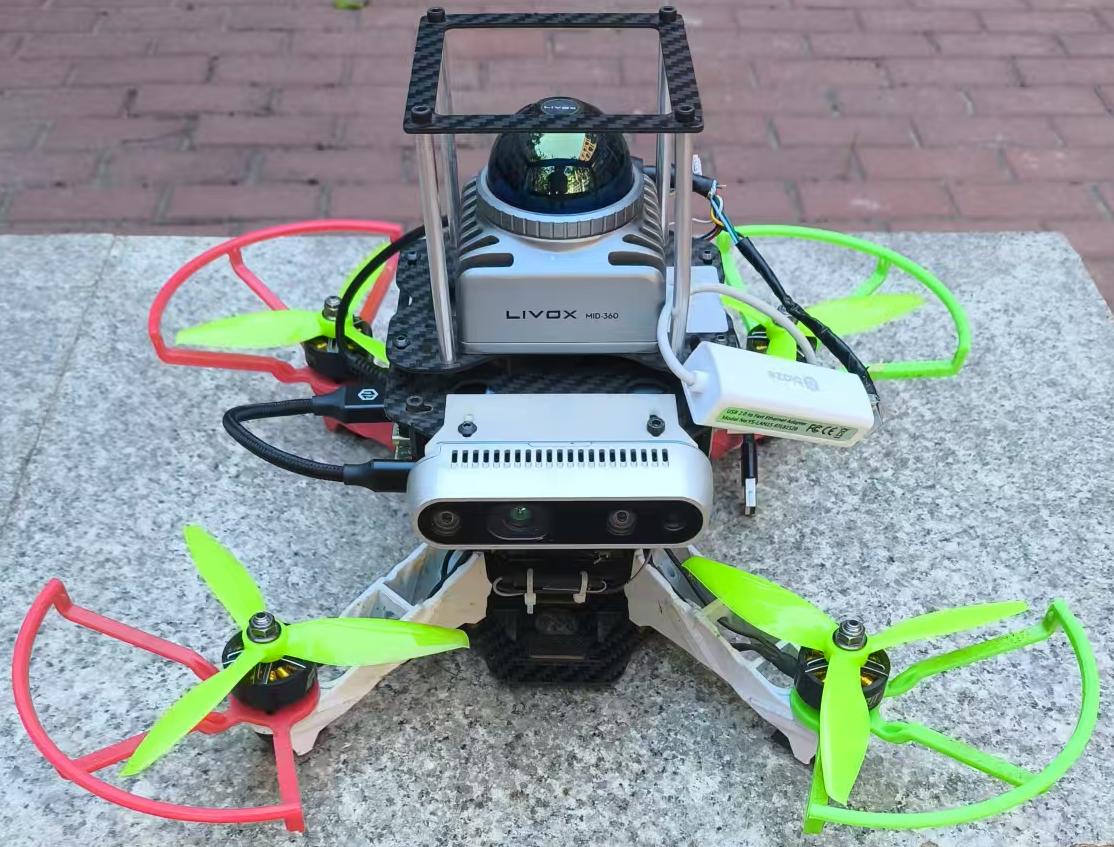}
\end{minipage}%
}
\subfloat[Courtyard.]{
\label{raw_vox_indoor}
\centering
\begin{minipage}[htp]{0.32\linewidth}
\centering
\includegraphics[width=1.0\textwidth]{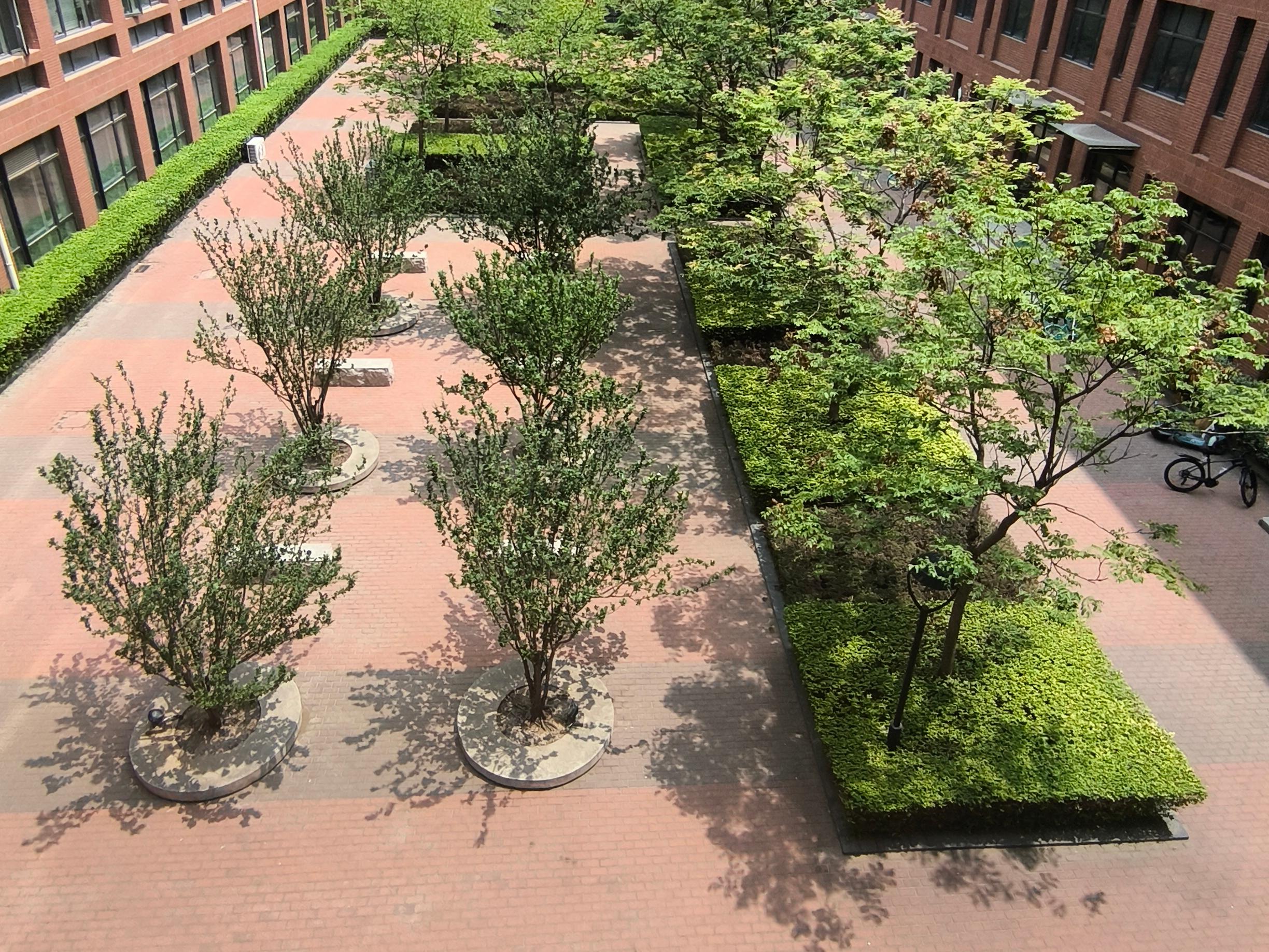}
\end{minipage}%
}
\subfloat[Park.]{
\label{raw_vox_indoor}
\centering
\begin{minipage}[htp]{0.323\linewidth}
\centering
\includegraphics[width=1.0\textwidth]{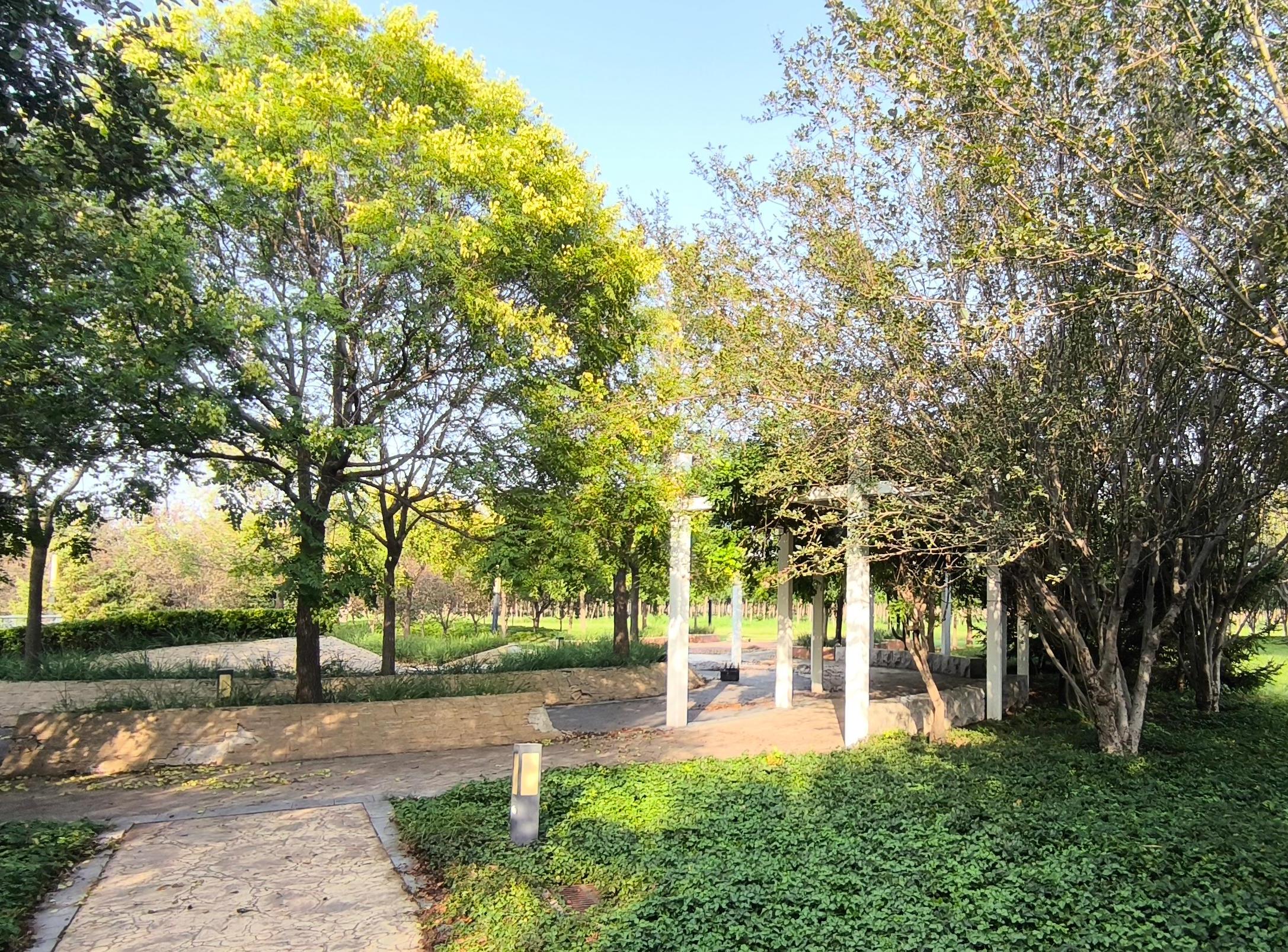}
\end{minipage}%
}\\
\subfloat[Mapping Result (Courtyard).]{
\label{raw_vox_indoor}
\centering
\begin{minipage}[htp]{0.53\linewidth}
\centering
\includegraphics[width=1.0\textwidth]{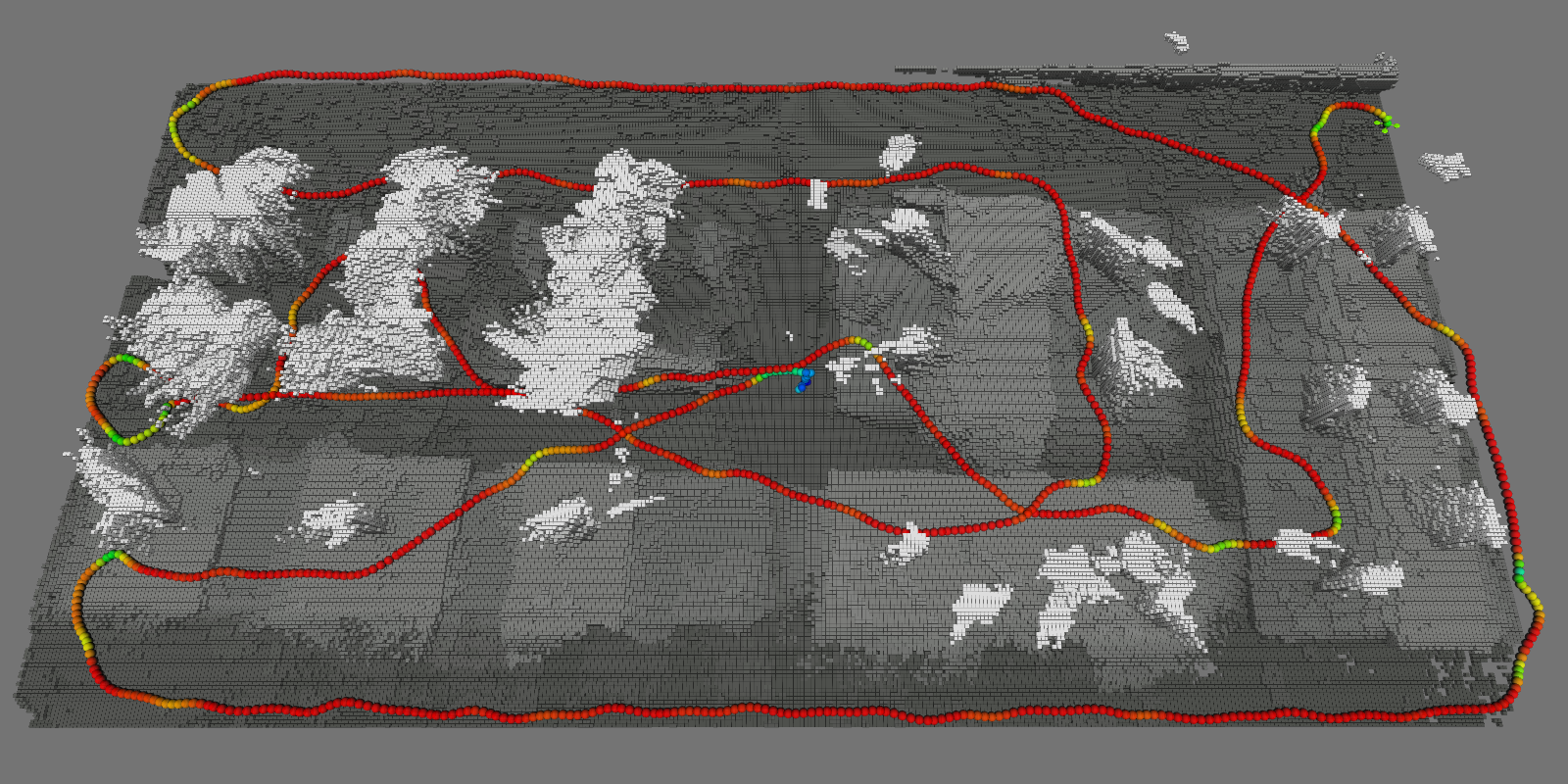}
\end{minipage}%
}
\subfloat[Mapping Result (Park).]{
\label{raw_vox_indoor}
\centering
\begin{minipage}[htp]{0.455\linewidth}
\centering
\includegraphics[width=1.0\textwidth]{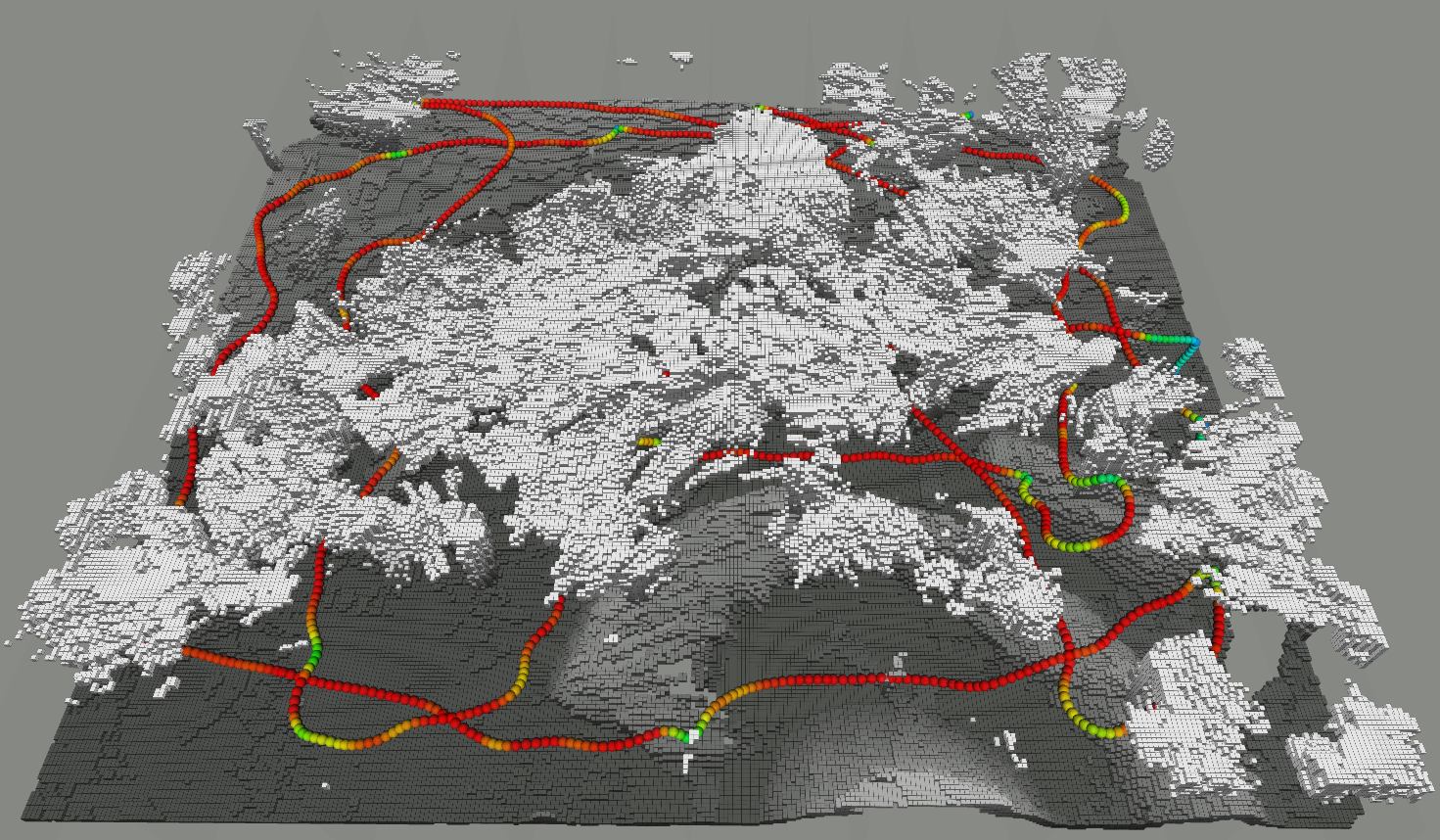}
\end{minipage}%
}\\
\subfloat[Explored Volume (Courtyard).]{
\label{raw_vox_indoor}
\centering
\begin{minipage}[htp]{0.485\linewidth}
\centering
\includegraphics[width=1.0\textwidth]{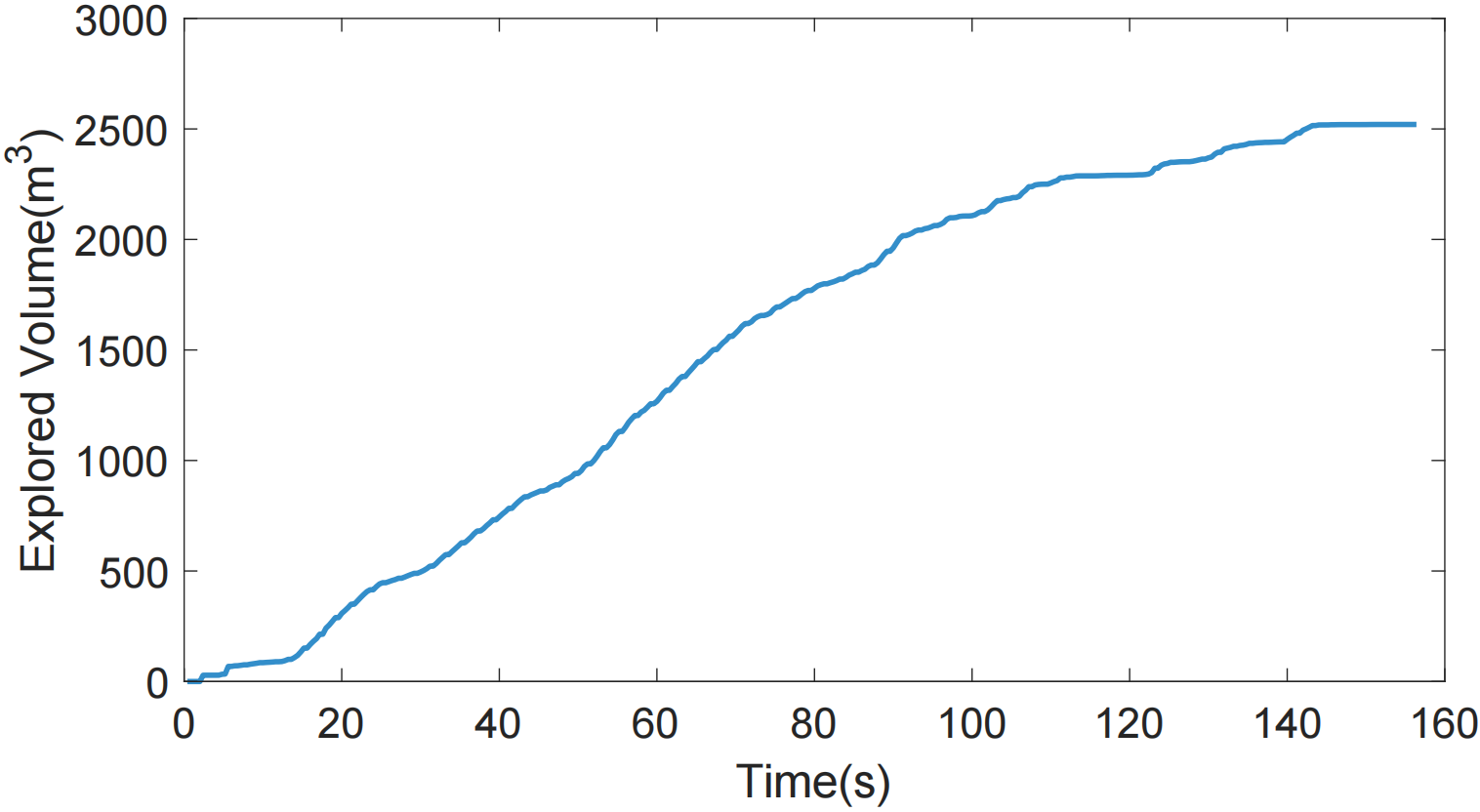}
\end{minipage}%
}
\subfloat[Explored Volume (Park).]{
\label{raw_vox_indoor}
\centering
\begin{minipage}[htp]{0.485\linewidth}
\centering
\includegraphics[width=1.0\textwidth]{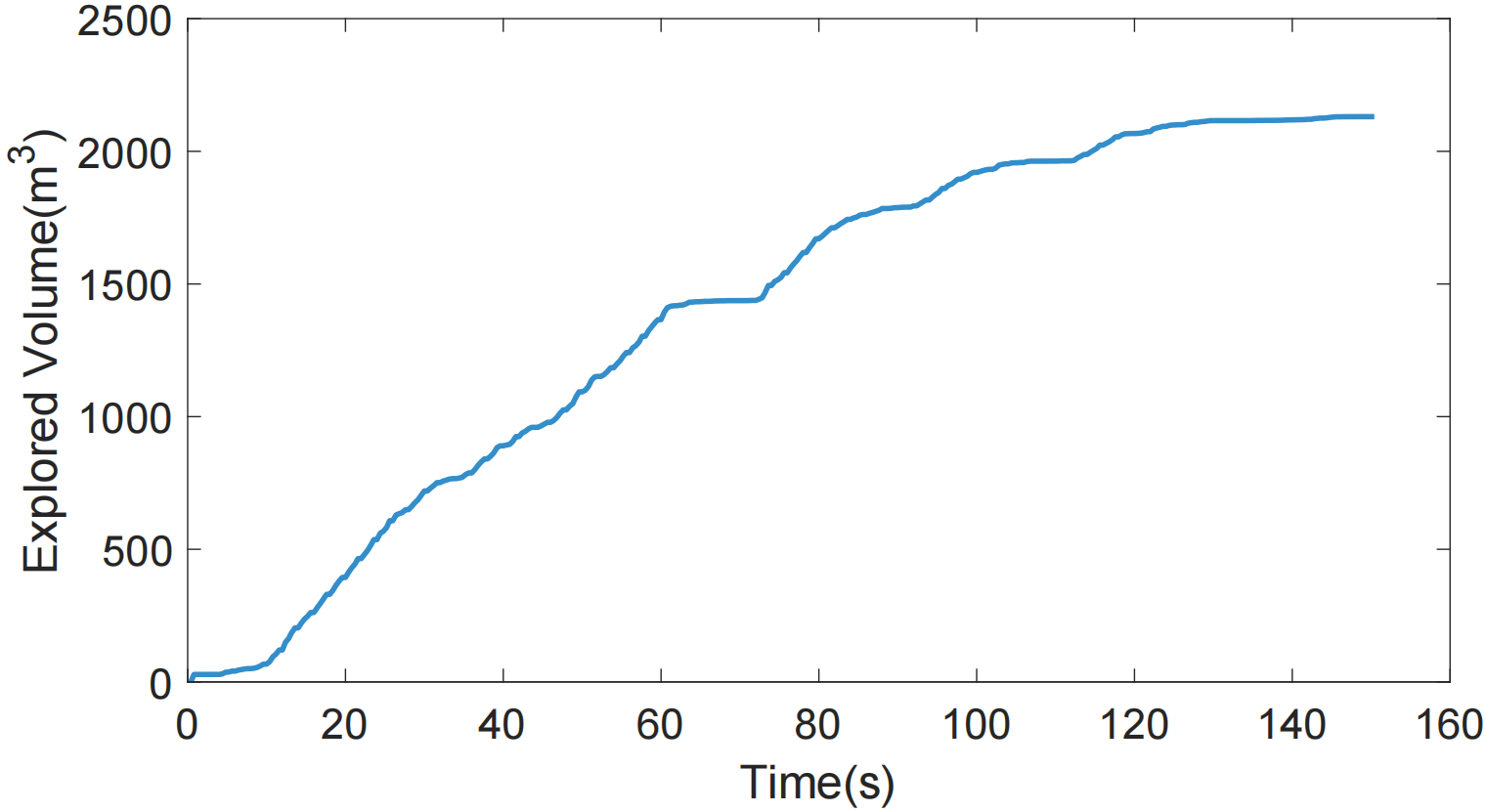}
\end{minipage}%
}\\
\caption{Experiment results of real-world explorations. }
\label{real}
\end{figure}

For comparative simulations, all the methods are implemented in C++ on a computer that runs the Ubuntu 20.04 LTS operation system, which uses Robot Operation System (ROS) for communication between the planner and simulator. The computer has an Intel Core i7-12700H, a GeForce 3060 12G, and 16G memory. For a fair comparison, all the methods are simulated on the platform that is provided by \cite{Yichen2024}. The FOV of the RGB-D camera used in the simulation is $[115,92]^\circ$, and its maximum sensing range is $5.0\,m$. As for real-world experiments, the proposed method is deployed on a self-made UAV platform (Fig.\ref{real} (a)). The proposed method is run on an on-board computer with an Intel Core i7-1260P. \Higher{The UAV is equipped with a Livox Mid 360 LiDAR for localization (using Point-LIO \cite{He_slam})} and an Intel Realsense D435i RGB-D camera for exploration. The FOV of the Realsense camera is $[87,58]^\circ$, and its maximum sensing range is $5.0\,m$. The parameters of our method are shown in Table. \ref{parameters}.

\begin{table}[htp]
	\caption{Parameters of the proposed method}  
	\label{parameters} 
		\resizebox{1.0\linewidth}{!}{
\begin{tabular}{cccc}  
\hline
	$r_v = 0.1m$ 							& $v_m=2.0\,m/s$           							& $a_m=3.0\,m/s^2$& $\dot{\psi}_m=1.57\,rad/s$	\\
	$\ddot{\psi}_m=1.57\,rad/s^2$  & $\lambda_a=0.6$  & $\lambda_e=0.3$&$\lambda_p=4000$  \\
	$\lambda_d=350$ & $\lambda_t=700$&\Higher{$r_u=0.35\,m$}&\Higher{$d_{th}=0.6\,m$}\\
	\Higher{$\psi_{th}=0.25\,rad$}&\Higher{$t_r=2.0\,s$}&\Higher{$t_o=0.1\,s$}&\\
\hline
	\end{tabular}}
\end{table}

\begin{figure*}[htp]
\centering
\subfloat[Proposed.]{
\label{raw_vox_cells}
\begin{minipage}[htp]{0.31\linewidth}
\centering
\includegraphics[width=1.0\textwidth]{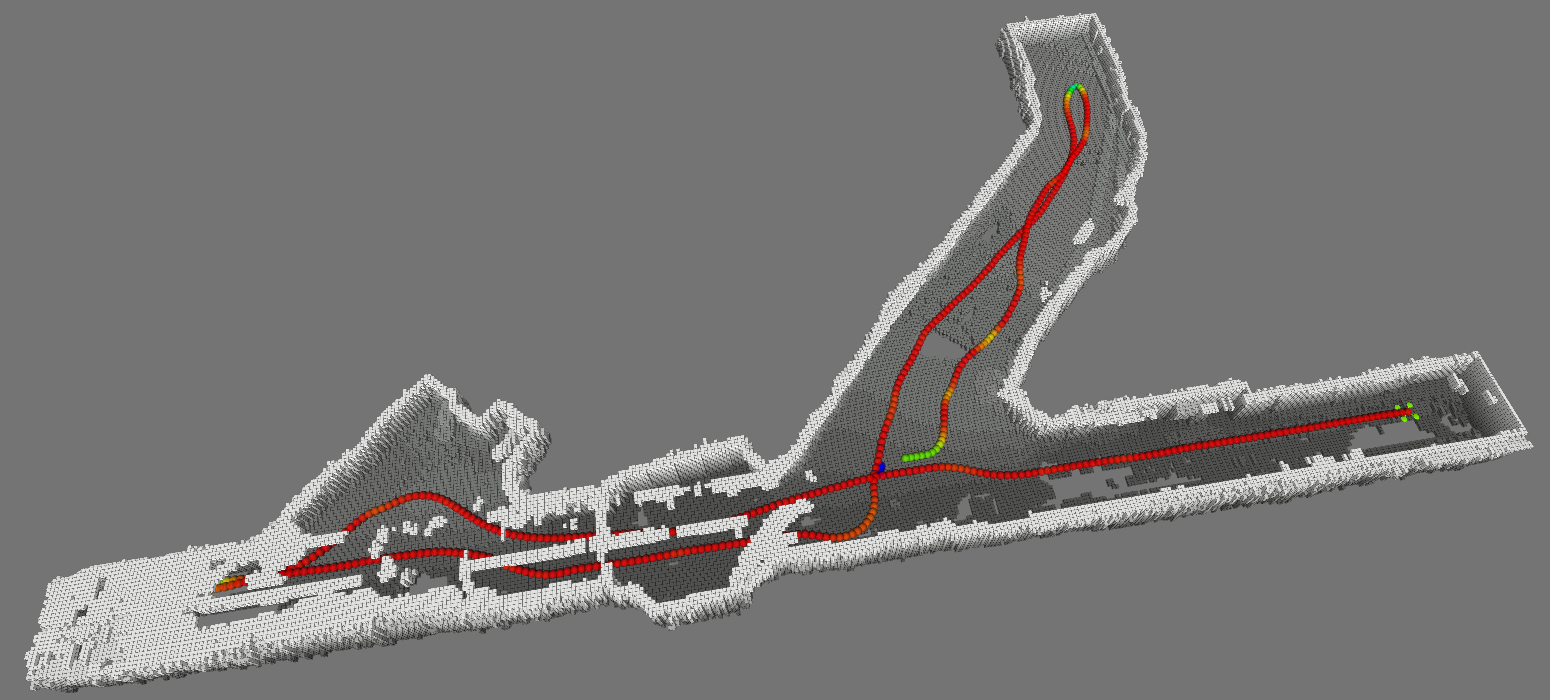}
\end{minipage}%
}
\subfloat[FALCON.]{
\label{raw_vox_indoor}
\centering
\begin{minipage}[htp]{0.31\linewidth}
\centering
\includegraphics[width=1.0\textwidth]{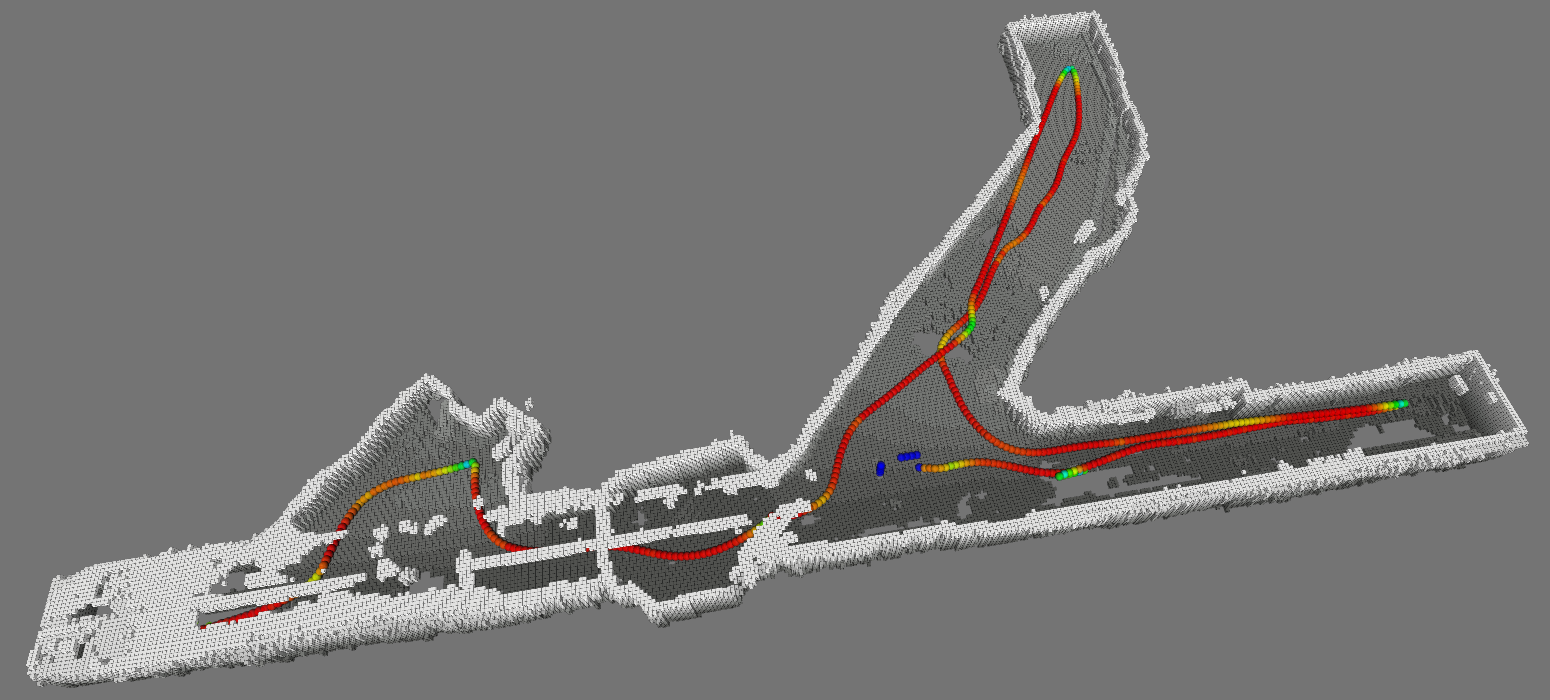}
\end{minipage}%
}%
\subfloat[FUEL.]{
\begin{minipage}[htp]{0.31\linewidth}
\centering
\includegraphics[width=1.0\textwidth]{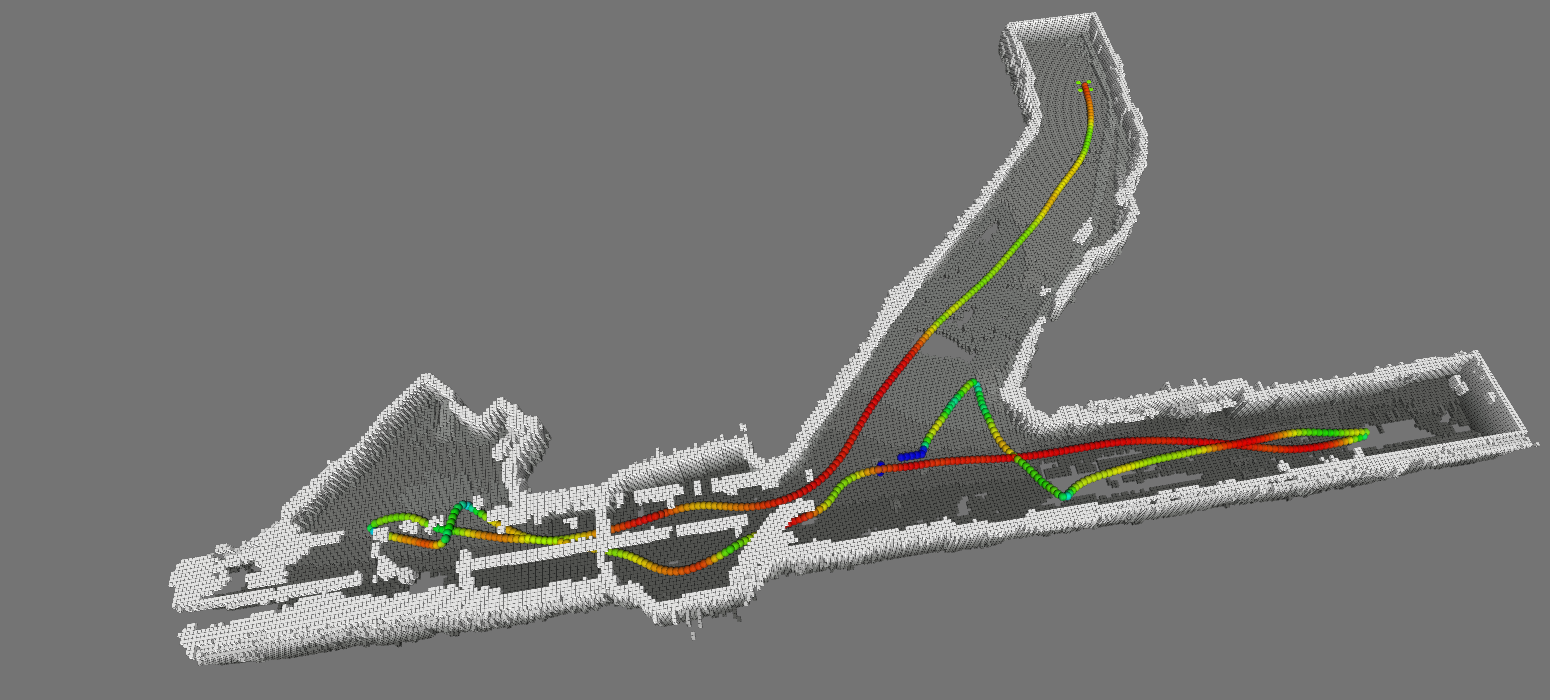}
\end{minipage}%
}\\
\subfloat[Proposed.]{
\label{raw_vox_cells}
\begin{minipage}[htp]{0.31\linewidth}
\centering
\includegraphics[width=1.0\textwidth]{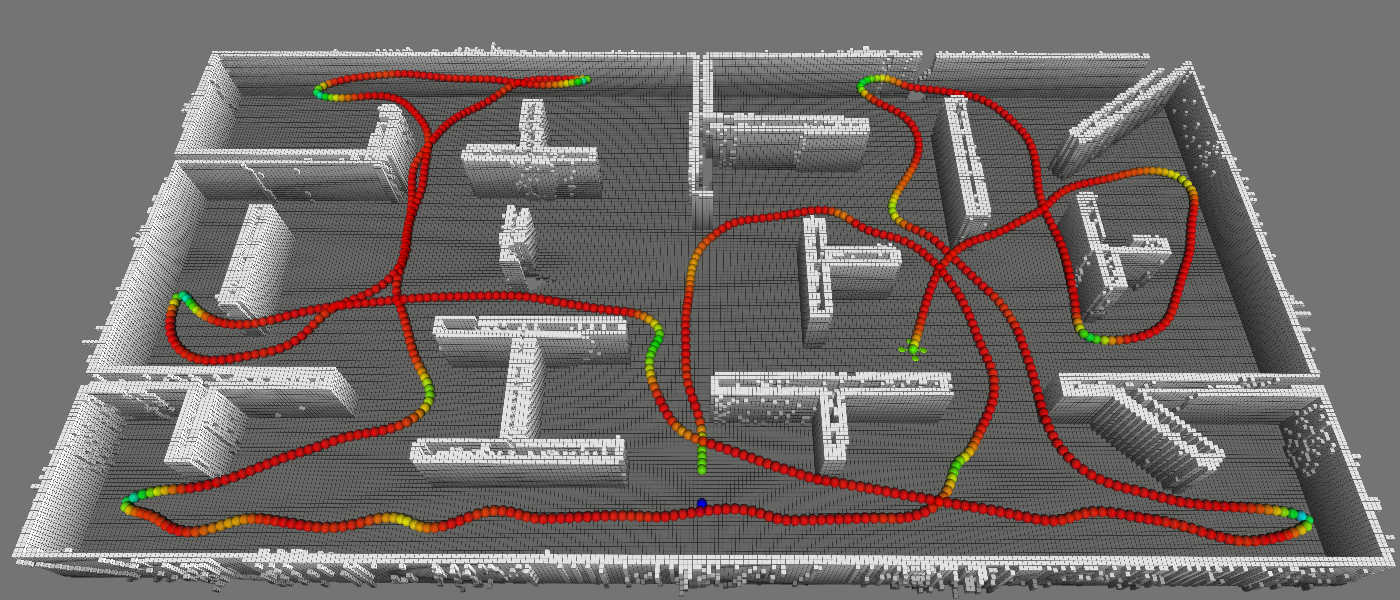}
\end{minipage}%
}
\subfloat[FALCON.]{
\label{raw_vox_indoor}
\centering
\begin{minipage}[htp]{0.31\linewidth}
\centering
\includegraphics[width=1.0\textwidth]{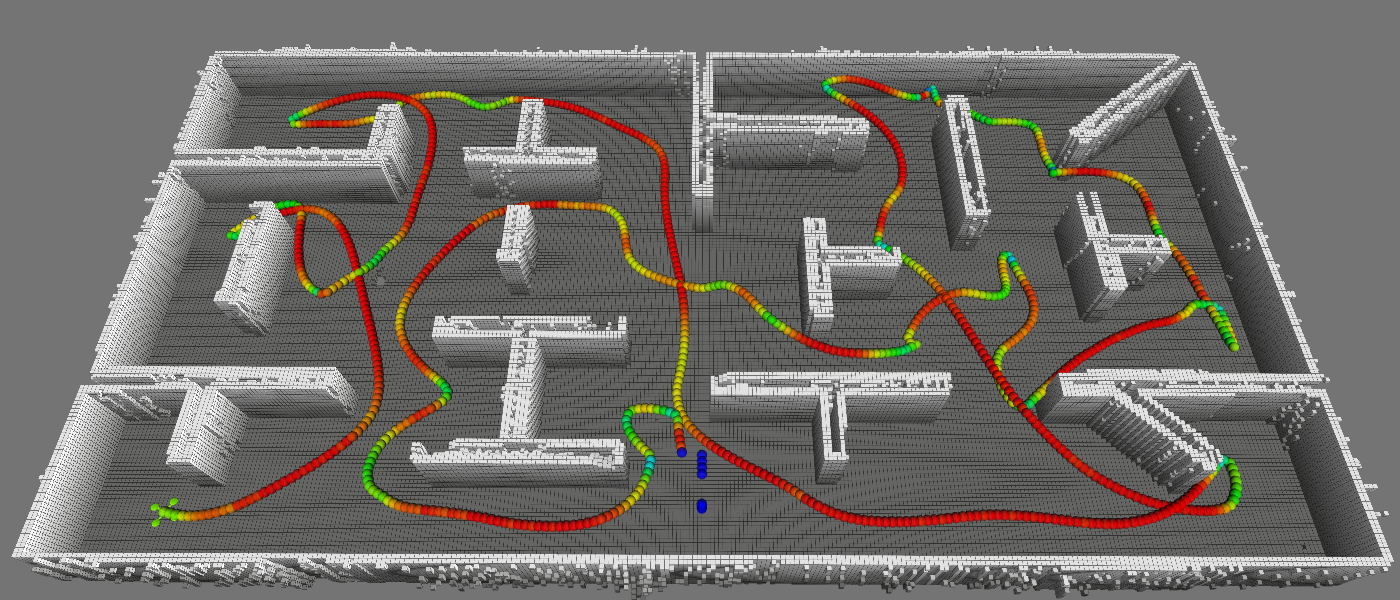}
\end{minipage}%
}%
\subfloat[FUEL.]{
\begin{minipage}[htp]{0.31\linewidth}
\centering
\includegraphics[width=1.0\textwidth]{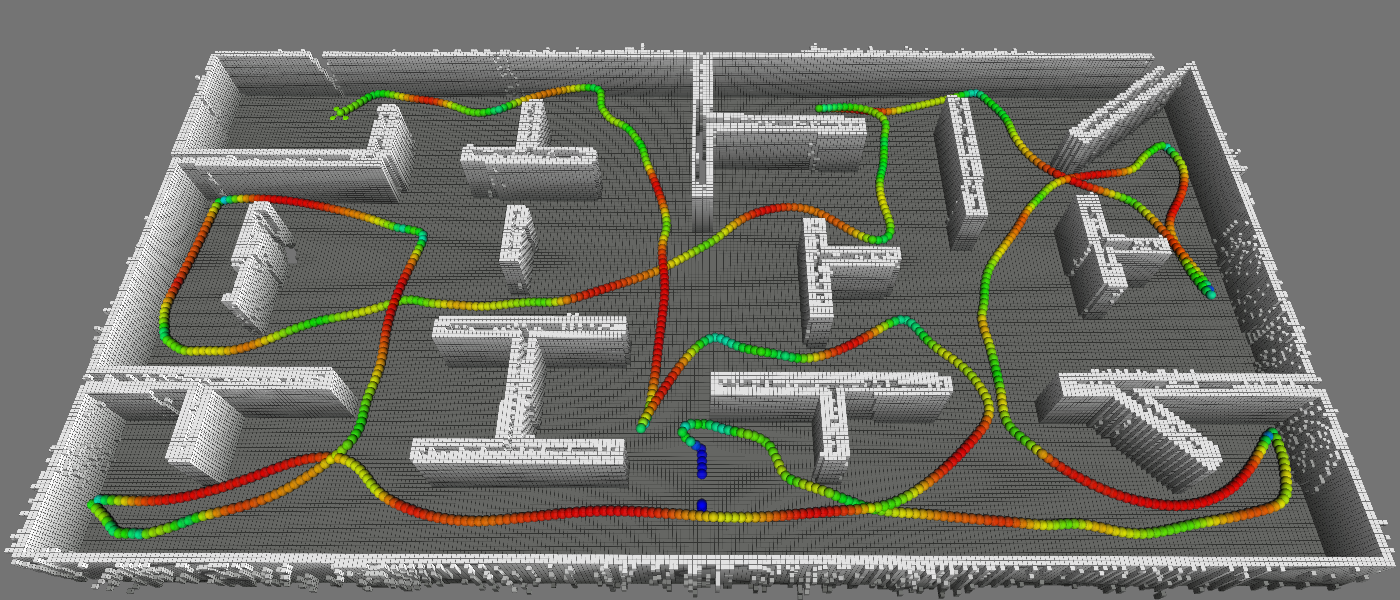}
\end{minipage}%
}\\
\subfloat[Proposed.]{
\label{raw_vox_cells}
\begin{minipage}[htp]{0.28\linewidth}
\centering
\includegraphics[width=1.0\textwidth]{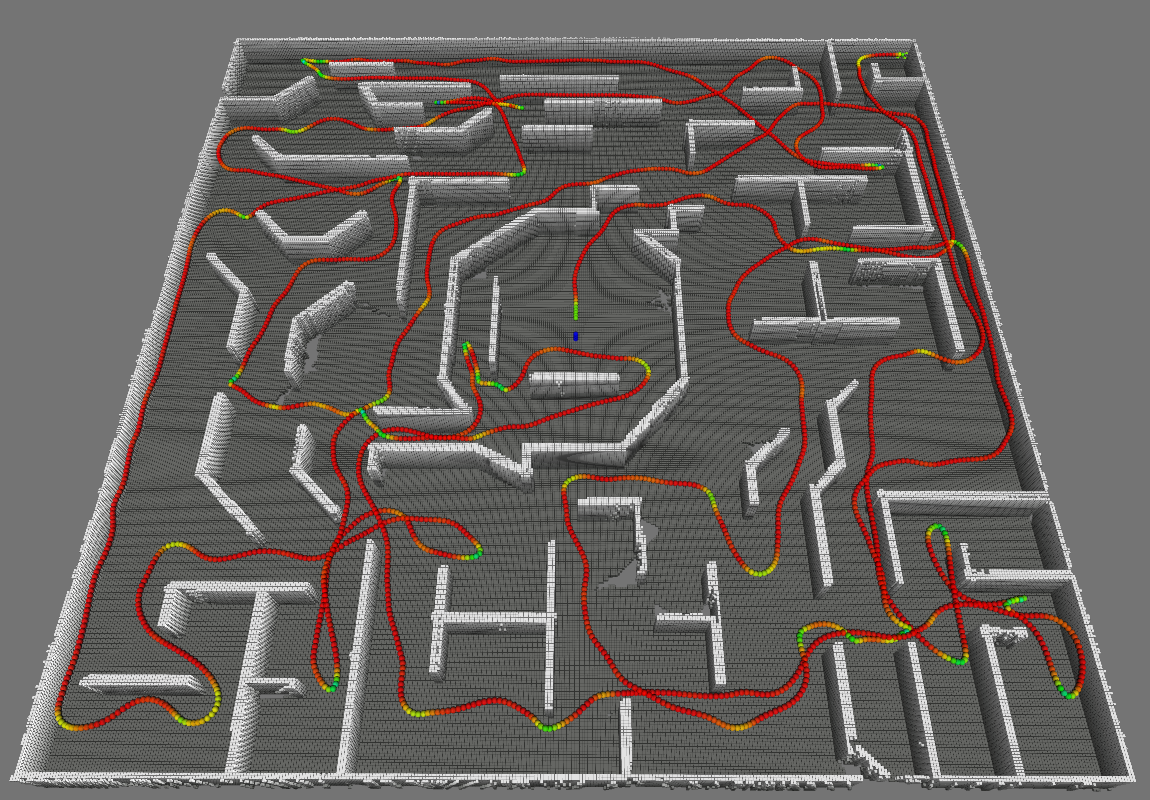}
\end{minipage}%
}
\subfloat[FALCON.]{
\label{raw_vox_indoor}
\centering
\begin{minipage}[htp]{0.28\linewidth}
\centering
\includegraphics[width=1.0\textwidth]{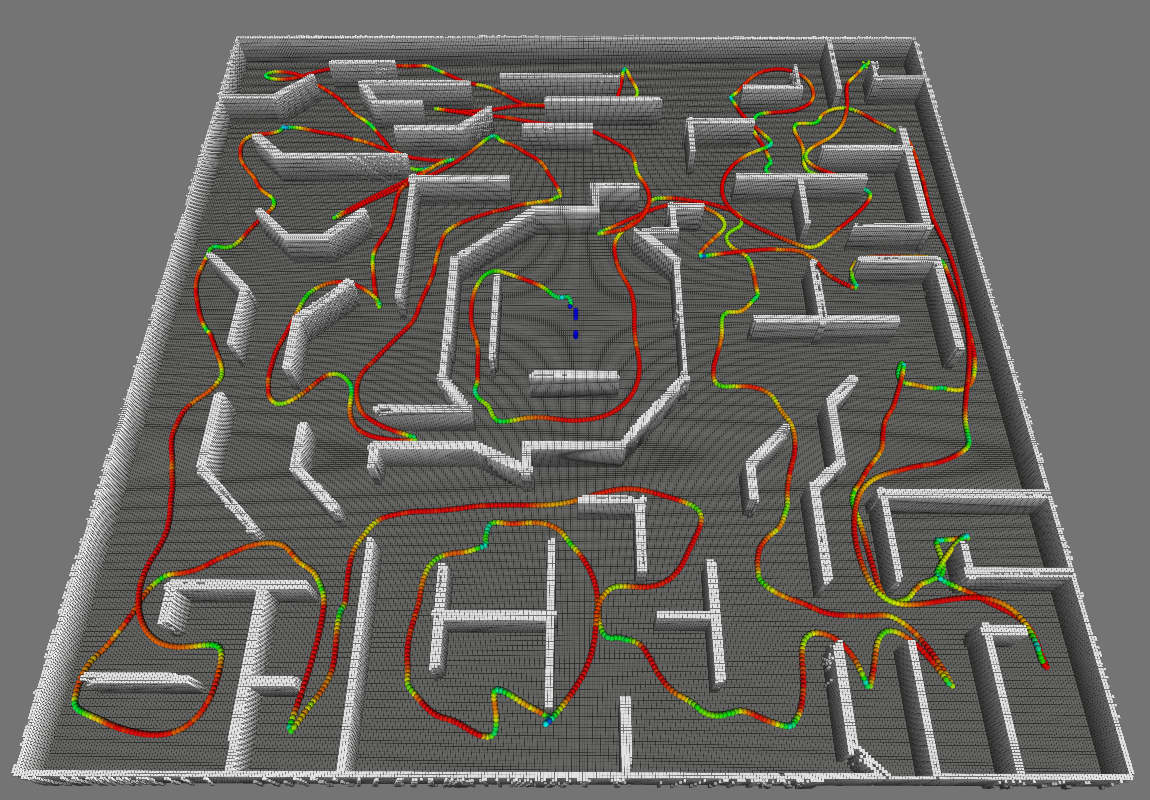}
\end{minipage}%
}%
\subfloat[FUEL.]{
\begin{minipage}[htp]{0.28\linewidth}
\centering
\includegraphics[width=1.0\textwidth]{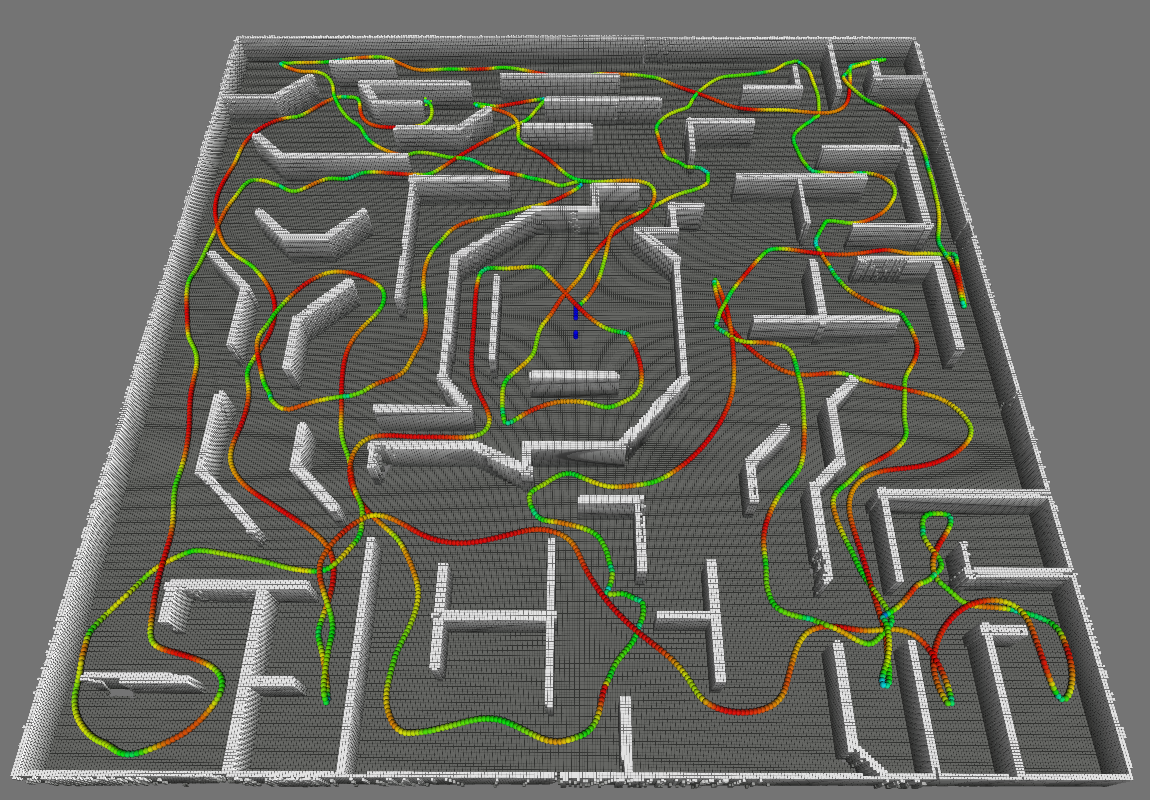}
\end{minipage}%
}
\subfloat[Velocity.]{
\centering
\begin{minipage}[htp]{0.08\linewidth}
\centering
\includegraphics[width=1.0\textwidth]{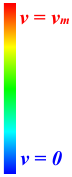}
\end{minipage}%
}\\
\subfloat[Proposed.]{
\label{raw_vox_cells}
\begin{minipage}[htp]{0.3\linewidth}
\centering
\includegraphics[width=1.0\textwidth]{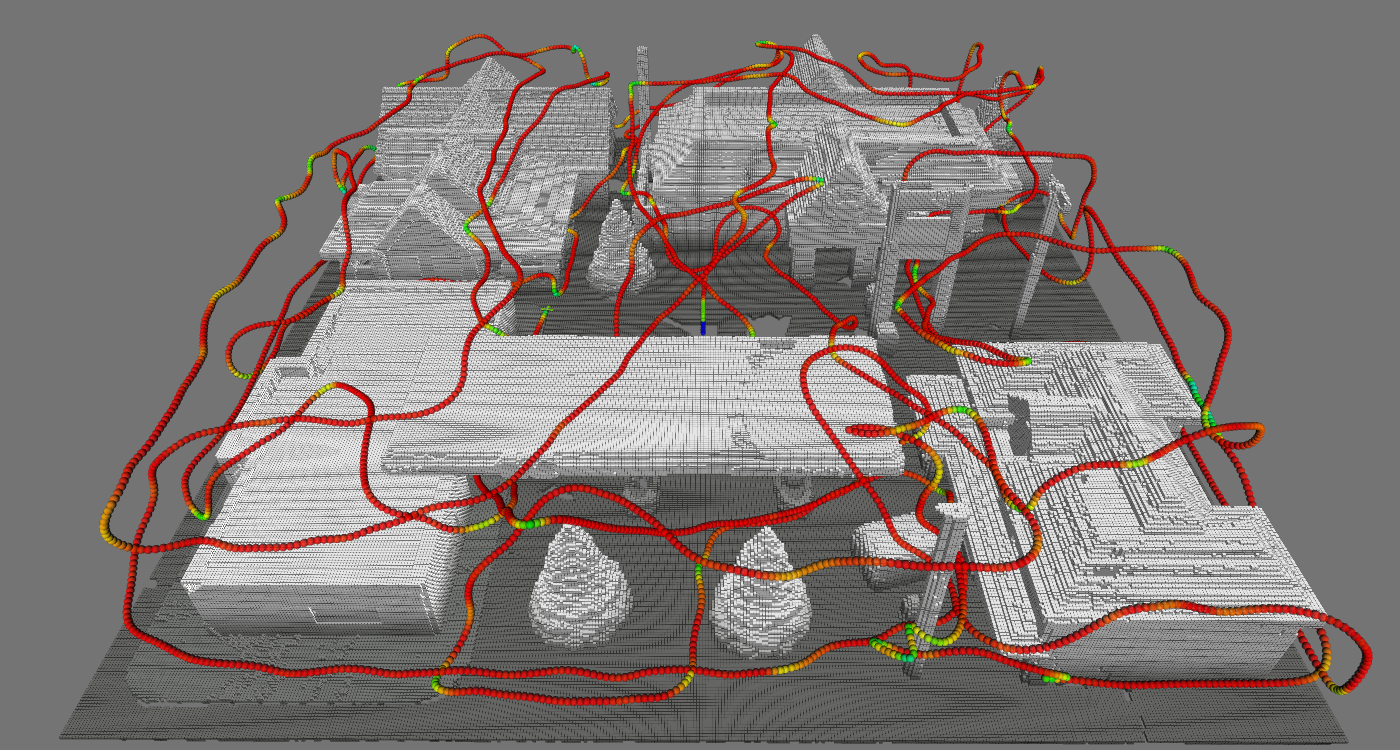}
\end{minipage}%
}
\subfloat[FALCON.]{
\label{raw_vox_indoor}
\centering
\begin{minipage}[htp]{0.3\linewidth}
\centering
\includegraphics[width=1.0\textwidth]{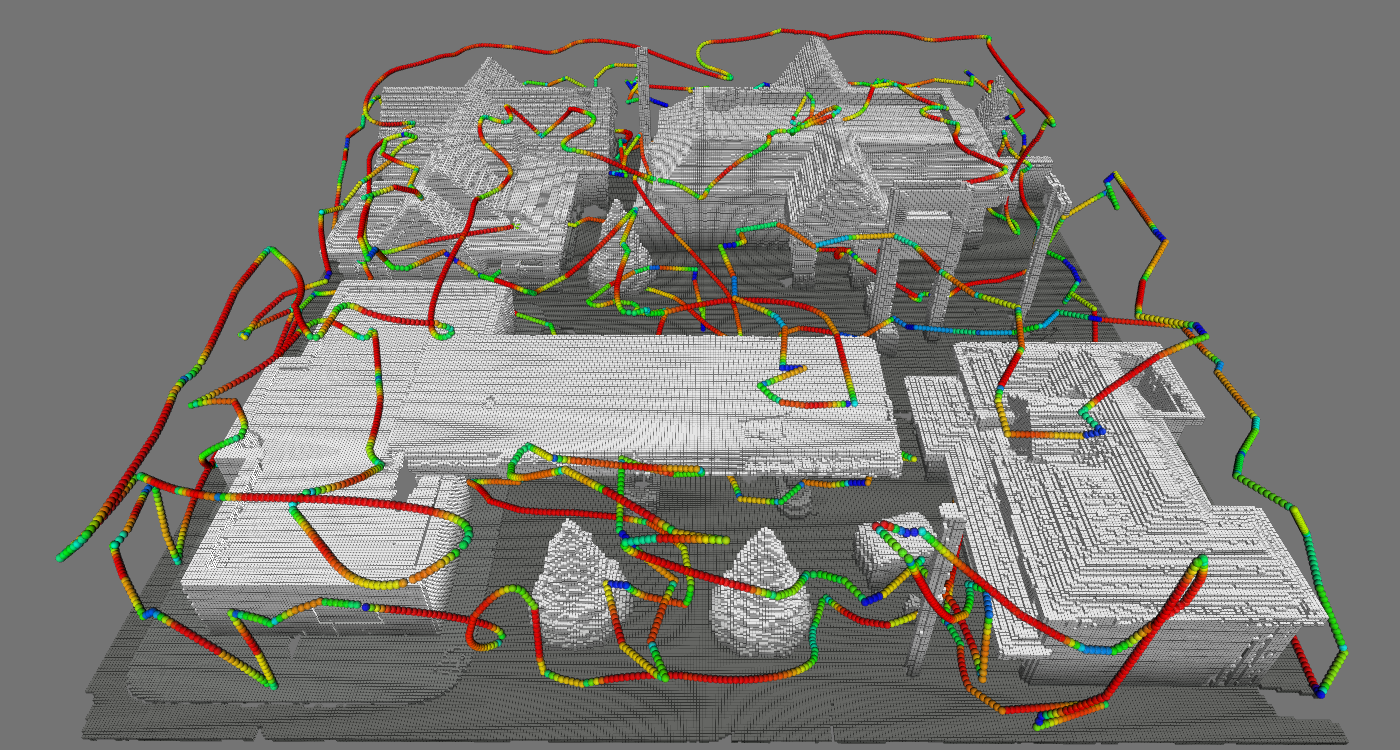}
\end{minipage}%
}%
\subfloat[FUEL.]{
\begin{minipage}[htp]{0.3\linewidth}
\centering
\includegraphics[width=1.0\textwidth]{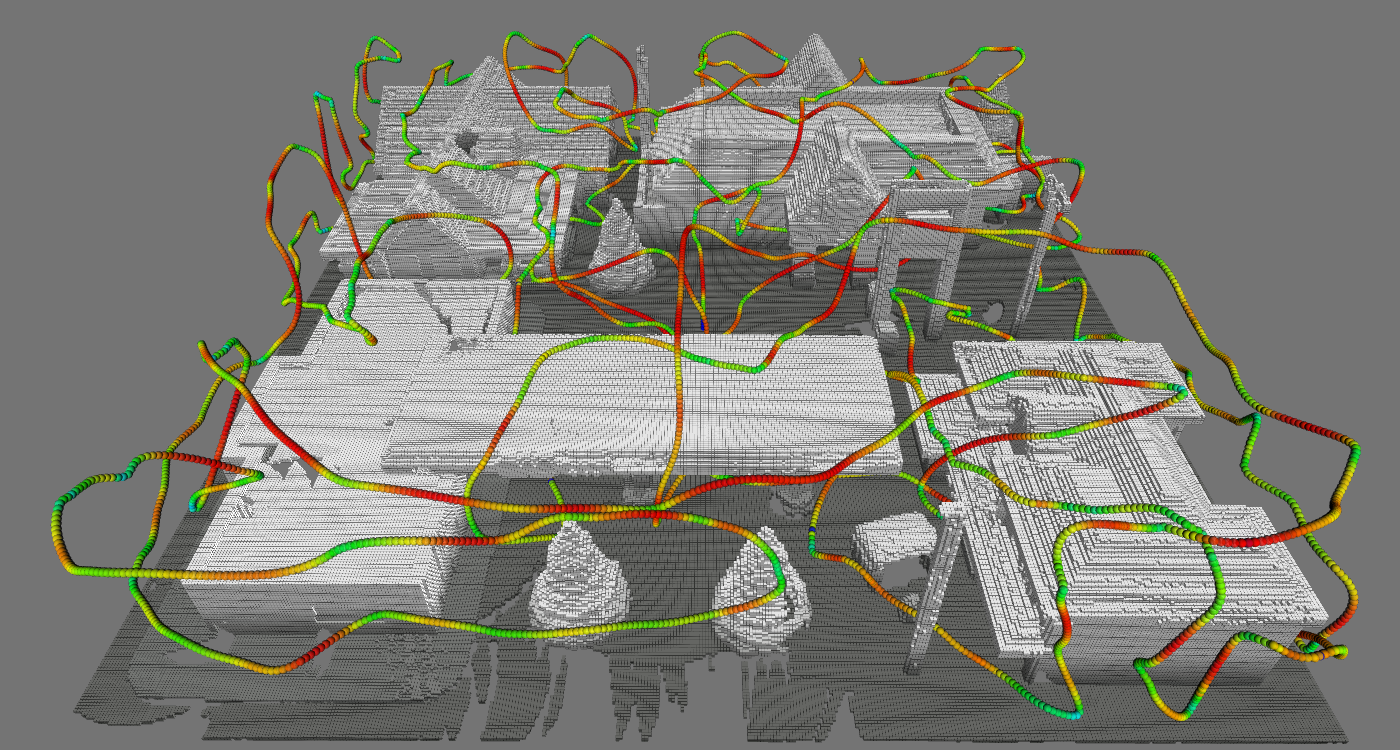}
\end{minipage}%
}\\
\subfloat[Explored Volume (DARPA Tunnel).]{
\centering
\begin{minipage}[htp]{0.235\linewidth}
\centering
\includegraphics[width=1.0\textwidth]{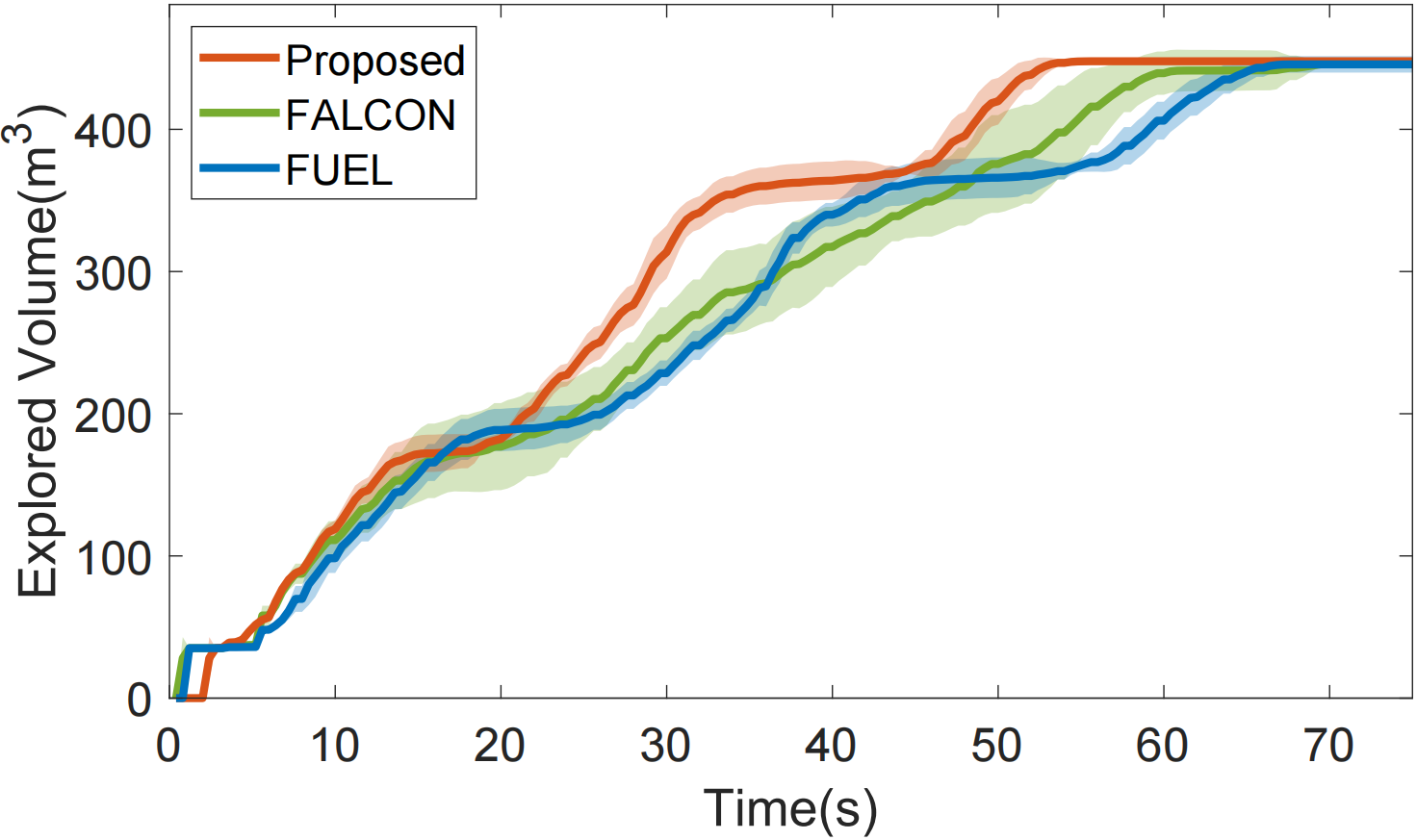}
\end{minipage}%
}
\subfloat[Explored Volume (Classical Office).]{
\centering
\begin{minipage}[htp]{0.235\linewidth}
\centering
\includegraphics[width=1.0\textwidth]{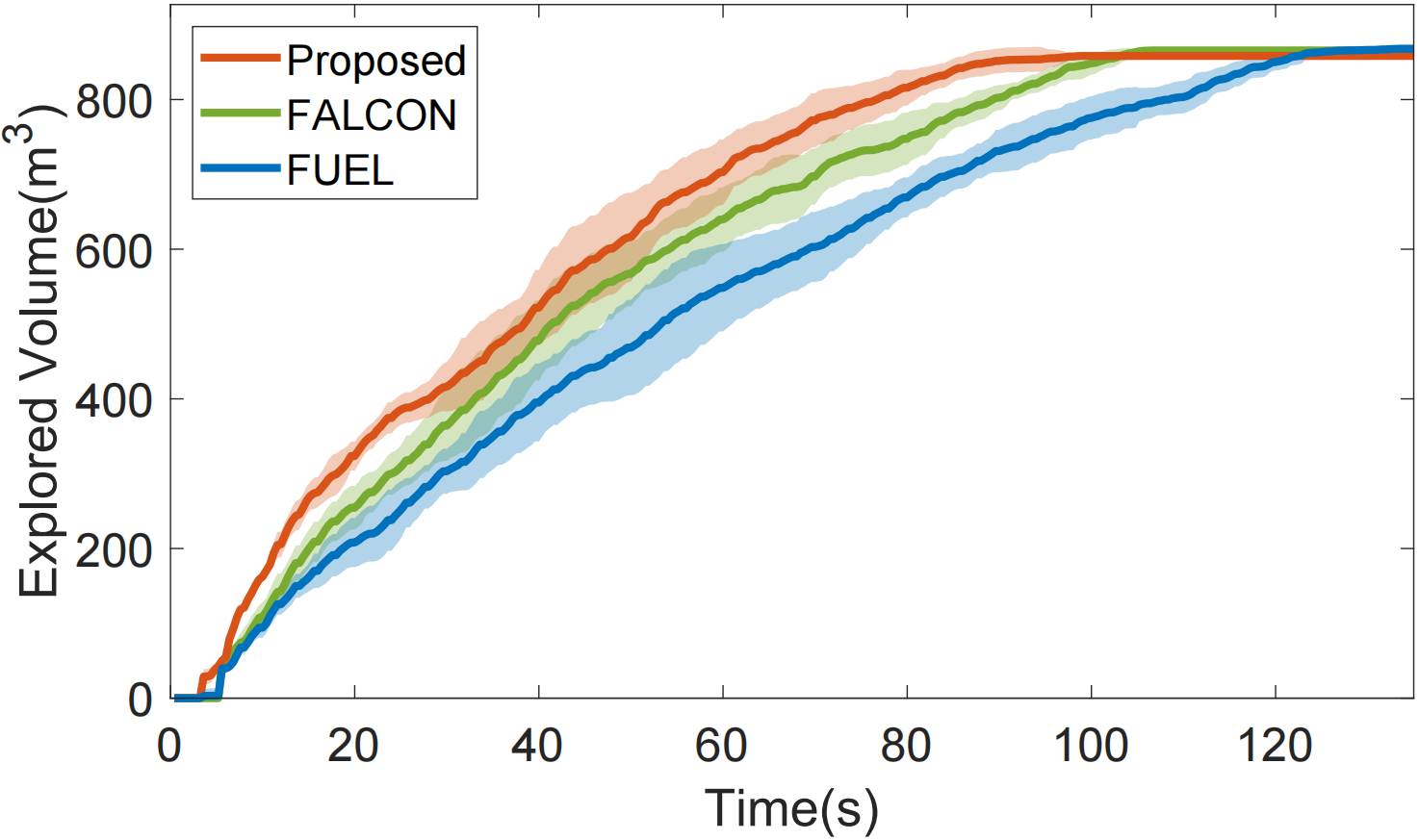}
\end{minipage}%
}
\subfloat[Explored Volume (Large Maze).]{
\centering
\begin{minipage}[htp]{0.245\linewidth}
\centering
\includegraphics[width=1.0\textwidth]{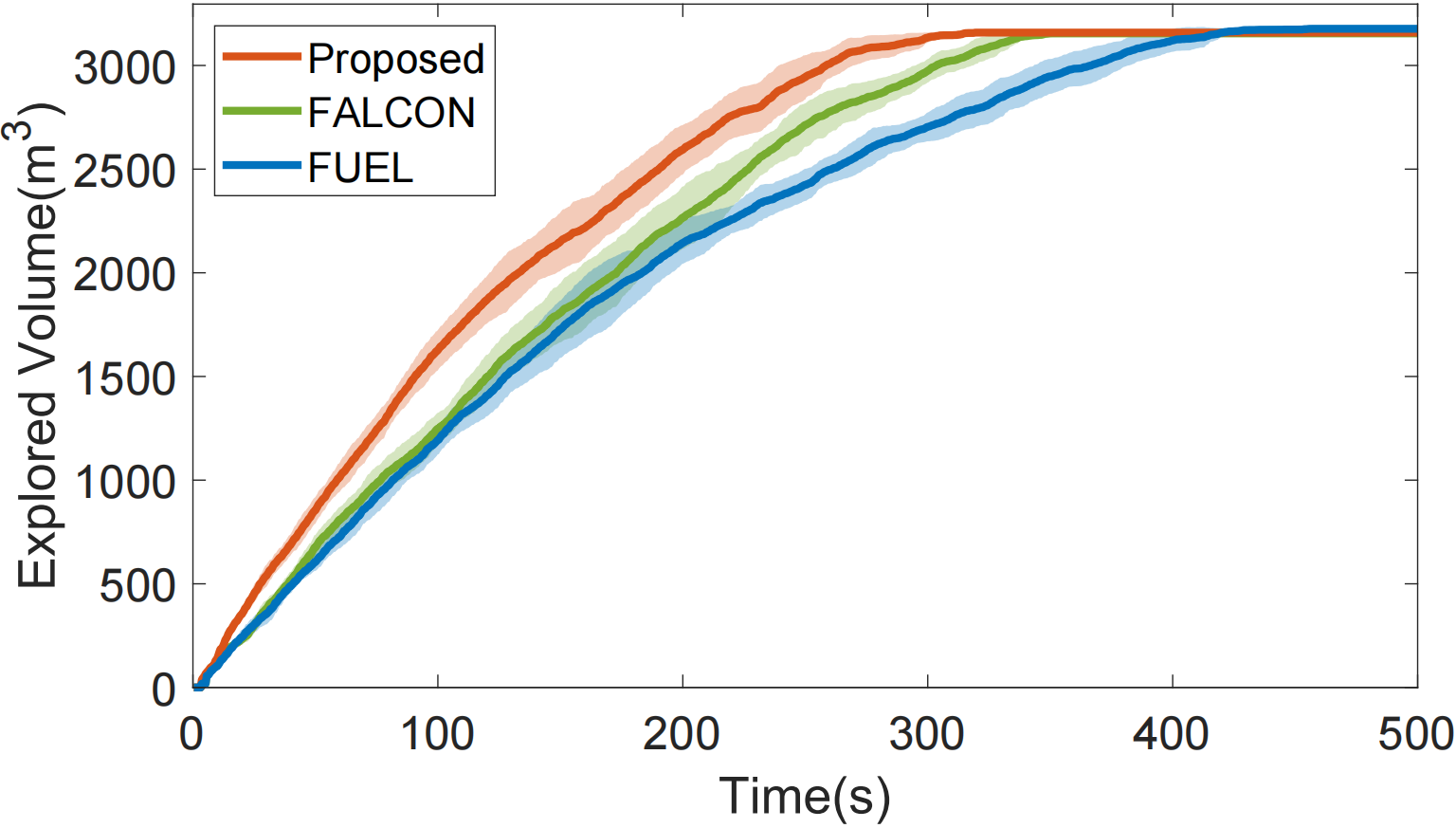}
\end{minipage}%
}
\subfloat[Explored Volume (City).]{
\centering
\begin{minipage}[htp]{0.25\linewidth}
\centering
\includegraphics[width=1.0\textwidth]{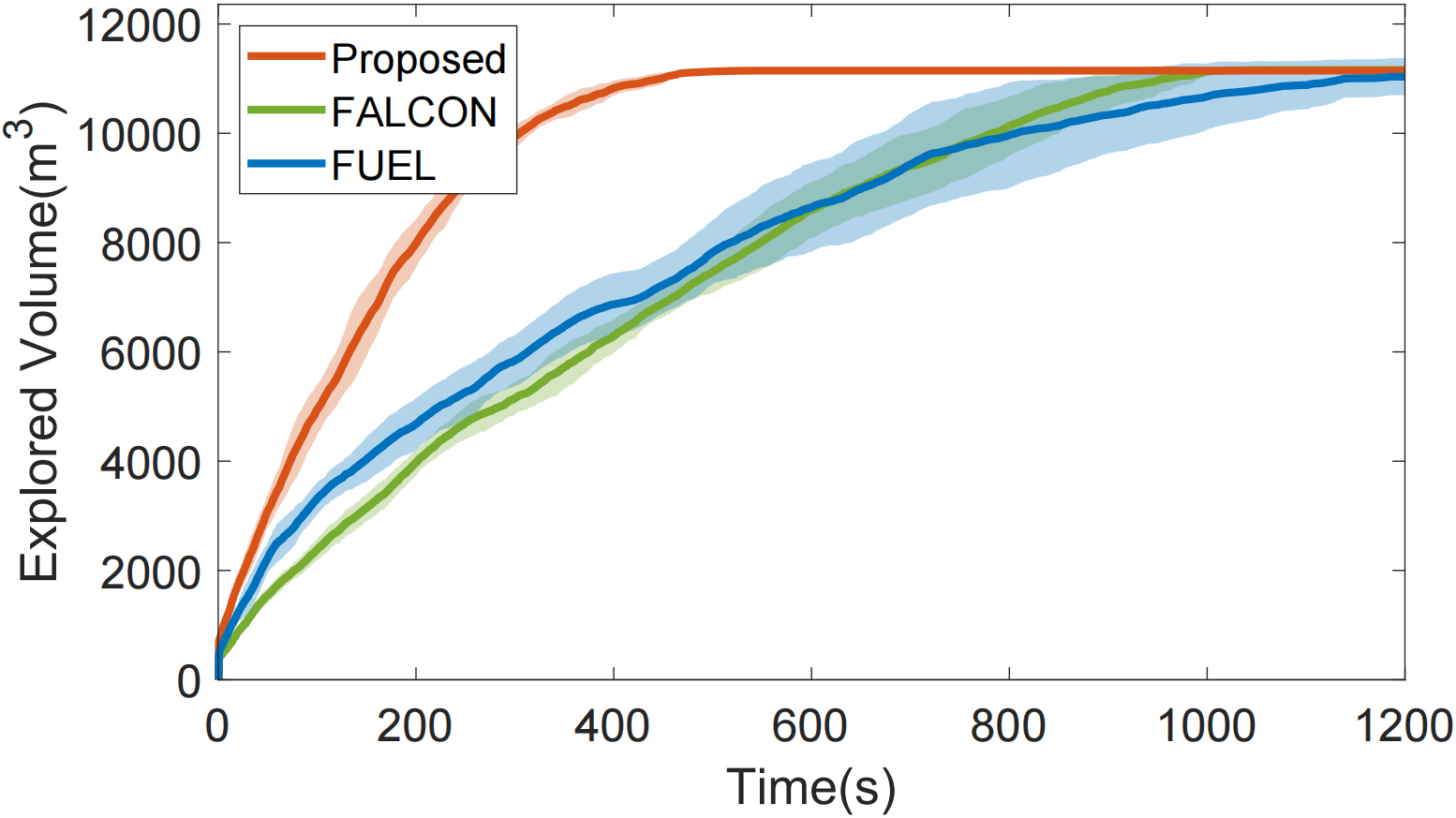}
\end{minipage}%
}\\
\caption{The mapping results and the executed trajectory of all the methods. The color of the trajectory (j) denotes the velocity of the UAV. Each row illustrates the simulations in one environment. The last row illustrates the explored volume vs. time.}
\label{sim}
\end{figure*}

\begin{table*}[tp]
	\centering  
	\caption{EXPLORATION STATISTIC OF ALL METHODS}  
	\label{metrics} 
	\resizebox{0.95\linewidth}{!}{
	\begin{tabular}{cccccccccc}  
	\toprule
		\textbf{Environment}&\textbf{Method}&\multicolumn{2}{c}{\textbf{Exploration Time} (s)}&\multicolumn{2}{c}{\textbf{Computational Cost} ($ms$)}&\multicolumn{2}{c}{\textbf{UAV Speed} ($m/s$)}&\multicolumn{2}{c}{\textbf{Traveled Distance} ($m$)}\\
		&&Avg&Std&Avg&Std&Avg&Std&Avg&Std\\
		\midrule
		\multirow{3}{*}{\textit{DARPA Tunnel}}&Proposed&{\textbf{53.48}}&\textbf{1.24}&{10.31}&\textbf{2.25}&\textbf{1.80}&\textbf{0.018}&{93.38}&\textbf{2.05}\\
		&FALCON&{56.24}&{4.50}&{19.80}&\textbf{6.13}&{1.63}&{0.039}&{96.88}&{5.84}\\
		&FUEL&{65.32}&{2.18}&\textbf{10.19}&{4.59}&{1.40}&{0.027}&\textbf{91.78}&{3.83}\\\cline{1-10}
		\multirow{3}{*}{\textit{Classical Office}}&Proposed&{\textbf{91.56}}&\textbf{3.60}&\textbf{14.42}&\textbf{2.94}&\textbf{1.79}&\textbf{0.016}&{160.86}&{6.64}\\
		&FALCON&{102.68}&{3.52}&{20.00}&{10.11}&{1.59}&{0.046}&\textbf{155.25}&\textbf{6.45}\\
		&FUEL&{128.56}&{5.71}&{22.76}&{18.74}&{1.31}&{0.045}&{167.89}&{10.65}\\\cline{1-10}
		\multirow{3}{*}{\textit{Large Maze}}&Proposed&{\textbf{301.64}}&\textbf{15.65}&\textbf{16.16}&\textbf{3.77}&\textbf{1.86}&{0.015}&{557.07}&{32.13}\\
		&FALCON&{343.84}&{12.45}&{63.38}&{33.03}&{1.61}&{0.018}&\textbf{554.98}&\textbf{21.64}\\
		&FUEL&{430.08}&{13.65}&{97.81}&{76.42}&{1.41}&\textbf{0.011}&{605.55}&{25.51}\\\cline{1-10}
		\multirow{3}{*}{\Higher{\textit{City}}}&\Higher{Proposed}&\Higher{\textbf{515.88}}&\Higher{\textbf{20.31}}&\Higher{\textbf{36.29}}&\Higher{\textbf{11.98}}&\Higher{\textbf{1.85}}&\Higher{\textbf{0.018}}&\Higher{\textbf{980.67}}&\Higher{{38.77}}\\
		&\Higher{FALCON}&\Higher{978.48}&\Higher{52.95}&\Higher{752.49}&\Higher{498.05}&\Higher{1.09}&\Higher{0.056}&\Higher{1083.57}&\Higher{\textbf{35.53}}\\
		&\Higher{FUEL}&\Higher{998.32}&\Higher{165.41}&\Higher{596.04}&\Higher{608.50}&\Higher{1.09}&\Higher{\textbf{0.173}}&\Higher{1081.11}&\Higher{100.53}\\\cline{1-10}
		\multirow{3}{*}{\textit{Large Tunnel}}&Proposed&{\textbf{2441.32}}&\textbf{51.78}&\textbf{20.48}&\textbf{5.70}&\textbf{3.32}&\textbf{0.017}&\textbf{8095.60}&\textbf{185.38}\\
		&FALCON&{-}&{-}&\textbf{-}&\textbf{-}&{-}&\textbf{-}&\textbf{-}&{-}\\
		&FUEL&{-}&{-}&{-}&{-}&{-}&{-}&-&-\\
\bottomrule
	\end{tabular}}
\end{table*}

\subsection{Simulation Results}
In this section, we compare the proposed method against \cite{Yichen2024, Boyu2021} in four simulation environments. \Higher{For fair comparison, we modify the velocity and acceleration optimizations of the comparative methods. The modification constrains their trajectory generation within the maximum linear velocity and acceleration, rather than within the three axes individually. Except for the size of the exploration space, the other parameters of the comparative methods are not modified.} Each method runs 10 times in a simulation environment. We record the exploration time, UAV speed, traveled distance, and computational cost for evaluation (shown in Table. \ref{metrics}).
\begin{itemize}
\item \textit{FALCON} \cite{Yichen2024}: This method plans the exploration sequence of unexplored regions by solving coverage-ATSP. After planning the exploration sequence, a time-optimal B-spline trajectory is generated utilizing \cite{Boyu2019}. 
\item \textit{FUEL} \cite{Boyu2021}: This method plans the exploration sequence of frontiers through solving frontier-ATSP. It also uses time-optimal B-spline trajectories for exploration. 
\end{itemize}

The first simulation environment is the $\textit{DARPA Tunnel}$ with a scale of $42\times20\times2\,m^3$. Each method runs $75\,s$. This environment is relatively simple, so the main factor that affects the exploration efficiency is the speed of the UAV. As shown in Fig.\ref{sim} (a-c), the proposed method maintains a higher speed than other methods ($10.4\%$ faster than FALCON and $28.6\%$ faster than FUEL). As a result, the proposed method is $5.2\%$ more efficient than FALCON and $16.1\%$ more efficient than FUEL. Fig. \ref{sim}(n) compares the explored volume of the three methods.

The second simulation environment is the $\textit{Classical Office}$ with a scale of $15\times30\times2\,m^3$. Each method runs $130\,s$. Compared with $\textit{DARPA Tunnel}$, $\textit{Classical Office}$ is more complex. The walls force the UAV to keep turning, which slows down the speed of the UAV. \Higher{Thanks to the curvature-penalized viewpoints planning and ASEO trajectory, the proposed method is able to reduce sharp turns and generate high-speed trajectories, while FALCON usually decelerates when meeting walls, and FUEL can reach high speed only in explored space, as shown in Fig. \ref{sim}(d-f). This is because the B-spline trajectories of FALCON and FUEL only try to minimize the current trajectory duration without considering continuous exploration. Moreover, the exploration efficiencies of FALCON and FUEL are limited because their implicit optimal assumptions are usually not fulfilled due to the wall obstacles and open space of this environment.} The UAV speed of the proposed method is $12.6\%$ faster than FALCON's and $36.6\%$ faster than FUEL's. Since this environment is more open than $\textit{DARPA Tunnel}$, the computational costs of the three methods grow. The computational cost of the proposed method is $29.3\%$ lower than FALCON's and $37.7\%$ lower than FUEL's. The traveled distances of the three methods are similar. As a result, the proposed method is $12.1\%$ more efficient than FALCON and $40.4\%$ more efficient than FUEL. Fig. \ref{sim}(o) compares the explored volume of the three methods.

The third simulation environment is the $\textit{Maze}$ with a scale of $40\times40\times2\,m^3$. Each method runs $500\,s$. \Higher{As shown by Fig. \ref{sim}(g-i), the proposed method keeps a high speed in this environment.} Because this environment is larger than previous ones, FALCON and FUEL have to spend more computing time on solving larger-scale ATSP. The computational cost of the proposed method also grows, but only $12.1\%$ compared to simulation in $\textit{Classical Office}$. This is because \Higher{EOHDT} can solve ATSP efficiently. \Higher{Similar to the result in $\textit{Classical Office}$, the traveled distance of the proposed method is slightly longer than FALCON's but shorter than FUEL's. This is caused by two factors. First, the curvature-penalized viewpoint planning makes the proposed method tend to choose to explore the target that can keep high UAV speed, which may leave some isolated unexplored corners and lead to the back-and-forth movements. Second, the routing solved by the proposed EOHDT is sub-optimal. Even so, the sacrifice of the traveled distance trade offers a lower computational cost and a faster UAV speed.}

\Higher{The fourth simulation environment is the $\textit{City}$ with a scale of $40\times40\times8.5\,m^3$. Each method runs $1200\,s$. Compared with previous environments, $\textit{City}$ has a larger size and contains denser obstacles, which pose significant challenges to the computational efficiency of each method. As shown by Fig. \ref{sim}(k-m), the proposed method executes a straighter trajectory and keeps a higher speed, while comparative methods' trajectories are more curved and their speeds are lower. The average computation time for both FALCON and FUEL exceeds $0.5\,s$, which has a significant impact on the exploration of real-time planning. In contrast, thanks to EOHDT, the proposed method is able to plan in real-time with massive boundary regions. Consequently, the proposed method is $89.7\%$ more efficient than FALCON and $93.5\%$ more efficient than FUEL. Moreover, the traveled distance of the proposed method is shorter than that of FALCON. This is because $\textit{City}$ contains dense building obstacles, which are far from FALCON's implicit optimal assumption.}

\begin{figure}[tp]
\centering
\subfloat[Executed Trajectory and Mapping Result.]{
\label{raw_vox_cells}
\begin{minipage}[htp]{0.905\linewidth}
\centering
\includegraphics[width=1.0\textwidth]{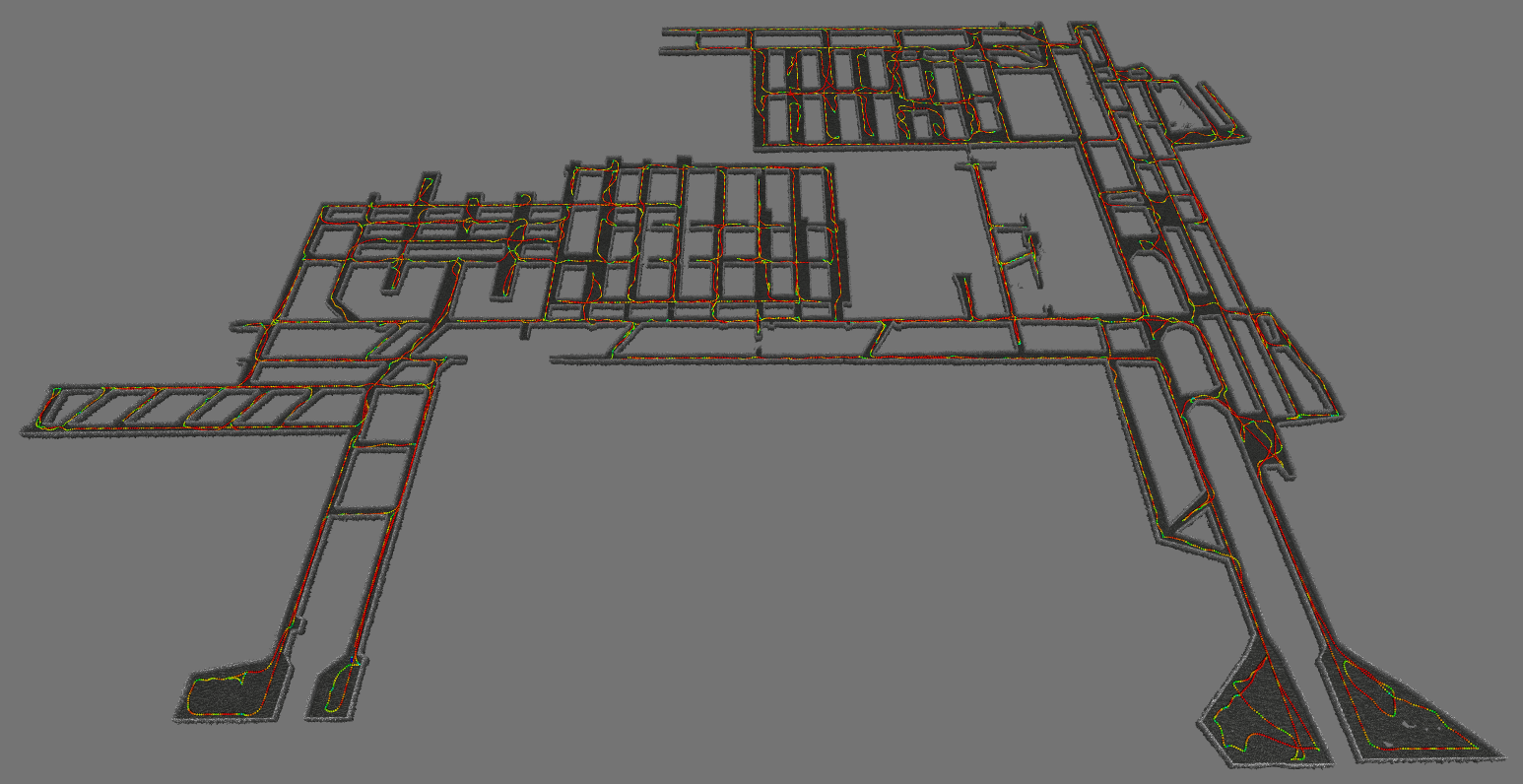}
\end{minipage}%
}\\
\subfloat[Explored Volume.]{
\label{raw_vox_indoor}
\centering
\begin{minipage}[htp]{0.88\linewidth}
\centering
\includegraphics[width=1.0\textwidth]{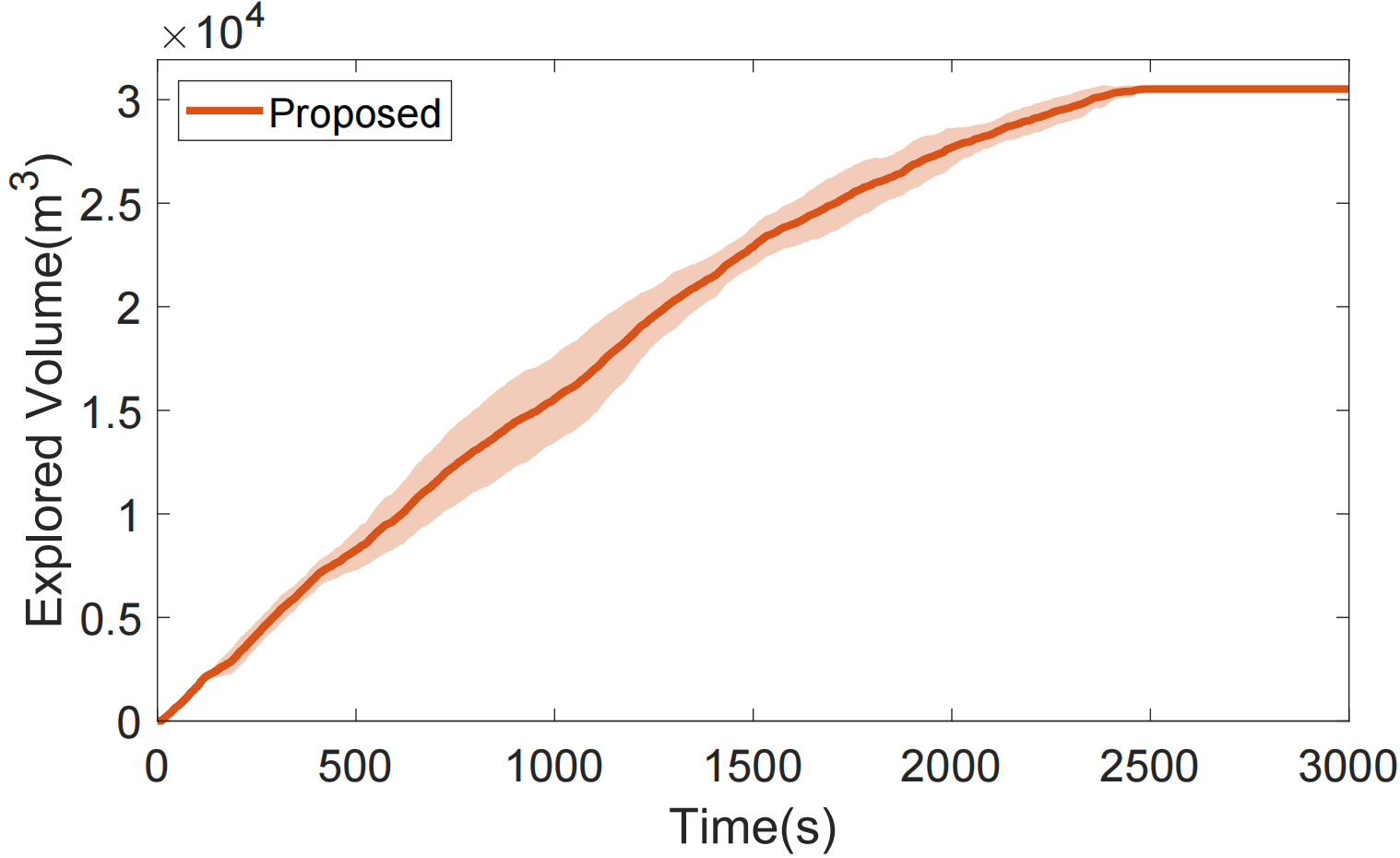}
\end{minipage}%
}%
\\
\caption{Simulation result in $\textit{Large Tunnel}$. }
\label{tunnel}
\end{figure}

The last simulation environment is the $\textit{Large Tunnel}$ with a scale of $251.5\times327.5\times2\,m^3$. Each method runs $3000\,s$. Since this environment is very large, we adopt $v_m=4.0m/s$ and $a_m=6.0m/s^2$ to save the simulation time. The purpose of this trial is to test the computational cost of the proposed method in an extremely large environment. \Higher{The comparative methods, FALCON and FUEL, fail to run in this environment due to their high memory usage and large computation.} On the contrary, the proposed method can accomplish the exploration as shown by Fig. \ref{tunnel}. The average computational cost of planning is $20.48\,ms$, which is low enough to plan in real time. The average speed of the UAV is $3.32\,m/s$. The explored volume of the proposed method is shown in Fig. \ref{tunnel}(b). The average exploration time is $2441.32\,s$. The result shows the proposed method is able to explore large-scale environments.

\subsection{Real-World Experiment}
\Higher{To evaluate the proposed method further, we deploy our method on a self-made UAV (Fig. \ref{real}(a)), and test it in two outdoor environments. The first environment is the $Courtyard$ with a scale of $42\times21\times3\,m^3$ as shown by Fig. \ref{real}(b). This environment is composed of trees and bushes. The exploration results are shown in Fig. \ref{real}(d) and Fig. \ref{real}(f). The exploration takes $145.1\,s$ to travel $260.9\,m$, explore $2520\,m^3$ of the environment. During the exploration, the average velocity of the UAV is $1.79\,m/s$, and the average planning time is $18.5\,ms$. The second environment is the $Park$ with a scale of $30\times25\times3\,m^3$ as shown by Fig. \ref{real}(c). This environment is composed of a pergola, uneven slopes, and cluttered trees, which is more challenging than the first environment. The exploration results are shown in Fig. \ref{real}(e) and Fig. \ref{real}(g). The exploration takes $145.5\,s$ to travel $256.4\,m$, explore $2131\,m^3$ of the environment. The average velocity of the UAV is $1.77\,m/s$, and the average planning time is $14.1\,ms$. Since this environment is more cluttered and smaller, the number of viewpoints to be evaluated and the number of nodes in ATSP are fewer, which reduces the UAV speed and the planning time. The results show the proposed method is able to plan a high-speed trajectory with limited computation resources online.}

The simulation results show that the proposed method can plan high-speed exploration trajectories in real-time, even in large 3-D environments. The real-world experiments show that the proposed method is also effective.

\section{CONCLUSION}
\Higher{In this work, we propose an efficient dual-layer exploration planning method (EDEN). By integrating the EOHDT algorithm for long-term region routing, the curvature-penalized viewpoint selection for short-term target planning, and the ASEO trajectory generation for high-speed continuous flight, EDEN significantly improves exploration performance.
Experimental results show that EDEN achieves up to $89.7\%$ higher exploration efficiency, $10.4\%-69.7\%$ faster UAV speed, and over $28.9\%$ reduction in computational cost compared to state-of-the-art methods. These advantages were consistently observed across multiple complex simulation environments and validated in real-world outdoor tests. The method enables real-time planning in large-scale environments while maintaining high-speed flight and safety. One drawback of the proposed method is that continuity-enhanced exploration planning increases the traveled distance. We plan to prioritize the exploration of isolated unknown regions to reduce redundant movements.}


\bibliographystyle{Bibliography/IEEEtranTIE}
\bibliography{bibFile}

\begin{thebibliography}{10}
\providecommand{\url}[1]{#1}
\csname url@samestyle\endcsname
\providecommand{\newblock}{\relax}
\providecommand{\bibinfo}[2]{#2}
\providecommand{\BIBentrySTDinterwordspacing}{\spaceskip=0pt\relax}
\providecommand{\BIBentryALTinterwordstretchfactor}{4}
\providecommand{\BIBentryALTinterwordspacing}{\spaceskip=\fontdimen2\font plus
\BIBentryALTinterwordstretchfactor\fontdimen3\font minus
  \fontdimen4\font\relax}
\providecommand{\BIBforeignlanguage}[2]{{%
\expandafter\ifx\csname l@#1\endcsname\relax
\typeout{** WARNING: IEEEtran.bst: No hyphenation pattern has been}%
\typeout{** loaded for the language `#1'. Using the pattern for}%
\typeout{** the default language instead.}%
\else
\language=\csname l@#1\endcsname
\fi
#2}}
\providecommand{\BIBdecl}{\relax}
\BIBdecl

\bibitem{zhang2022fast}
S.~Zhang, X.~Zhang, T.~Li, J.~Yuan, and Y.~Fang, ``Fast active aerial
  exploration for traversable path finding of ground robots in unknown
  environments,'' \emph{IEEE Transactions on Instrumentation and Measurement},
  vol.~71, pp. 1--13, 2022.

\bibitem{Shuang2024}
S.~Qi, B.~Lin, Y.~Deng, X.~Chen, and Y.~Fang, ``Minimizing maximum latency of
  task offloading for multi-{UAV}-assisted maritime search and rescue,''
  \emph{IEEE Transactions on Vehicular Technology}, vol.~73, no.~9, pp.
  13\,625--13\,638, 2024.

\bibitem{Yao2024}
H.~Yao and X.~Liang, ``Autonomous exploration under canopy for forest
  investigation using lidar and quadrotor,'' \emph{IEEE Transactions on
  Geoscience and Remote Sensing}, vol.~62, pp. 1--19, 2024.

\bibitem{Shuaizheng2023}
S.~Yan, Z.~Wu, J.~Wang, Y.~Huang, M.~Tan, and J.~Yu, ``Real-world learning
  control for autonomous exploration of a biomimetic robotic shark,''
  \emph{IEEE Transactions on Industrial Electronics}, vol.~70, no.~4, pp.
  3966--3974, 2023.

\bibitem{Jie2023}
J.~Liu, Y.~Lv, Y.~Yuan, W.~Chi, G.~Chen, and L.~Sun, ``An efficient robot
  exploration method based on heuristics biased sampling,'' \emph{IEEE
  Transactions on Industrial Electronics}, vol.~70, no.~7, pp. 7102--7112,
  2023.

\bibitem{Feng2024}
C.~Feng, H.~Li, M.~Zhang, X.~Chen, B.~Zhou, and S.~Shen, ``{FC-Planner}: A
  skeleton-guided planning framework for fast aerial coverage of complex {3D}
  scenes,'' in \emph{2024 IEEE International Conference on Robotics and
  Automation (ICRA)}, pp. 8686--8692, 2024.

\bibitem{Xuetao2024}
X.~Zhang, X.~Xu, Y.~Liu, H.~Wang, X.~Zhang, and Y.~Zhuang, ``{FGIP}: A
  frontier-guided informative planner for {UAV} exploration and
  reconstruction,'' \emph{IEEE Transactions on Industrial Informatics},
  vol.~20, no.~4, pp. 6155--6166, 2024.

\bibitem{Mingjie2024}
M.~Zhang, C.~Feng, Z.~Li, G.~Zheng, Y.~Luo, Z.~Wang, J.~Zhou, S.~Shen, and
  B.~Zhou, ``{SOAR}: Simultaneous exploration and photographing with
  heterogeneous {UAVs} for fast autonomous reconstruction,'' in \emph{2024
  IEEE/RSJ International Conference on Intelligent Robots and Systems (IROS)},
  pp. 10\,975--10\,982, 2024.

\bibitem{xuetao2022}
X.~Zhang, Y.~Chu, Y.~Liu, X.~Zhang, and Y.~Zhuang, ``A novel informative
  autonomous exploration strategy with uniform sampling for quadrotors,''
  \emph{IEEE Transactions on Industrial Electronics}, vol.~69, no.~12, pp.
  13\,131--13\,140, 2022.

\bibitem{Bai2024}
X.~Bai, C.~Li, B.~Zhang, Z.~Wu, and S.~S. Ge, ``Efficient performance impact
  algorithms for multirobot task assignment with deadlines,'' \emph{IEEE
  Transactions on Industrial Electronics}, vol.~71, no.~11, pp.
  14\,373--14\,382, 2024.

\bibitem{Bai2023}
X.~Bai, C.~Li, C.~Li, A.~Khan, T.~Zhang, and B.~Zhang, ``Multi-robot task
  assignment for serving people quarantined in multiple hotels during
  {COVID-19} pandemic,'' \emph{Quantitative Imaging in Medicine and Surgery},
  vol.~13, no.~3, p. 1802, 2023.

\bibitem{Yamauchi1997}
B.~Yamauchi, ``A frontier-based approach for autonomous exploration,'' in
  \emph{Proceedings 1997 IEEE International Symposium on Computational
  Intelligence in Robotics and Automation CIRA'97. 'Towards New Computational
  Principles for Robotics and Automation'}, pp. 146--151, 1997.

\bibitem{Cieslewski2017}
T.~Cieslewski, E.~Kaufmann, and D.~Scaramuzza, ``Rapid exploration with
  multi-rotors: A frontier selection method for high speed flight,'' in
  \emph{2017 IEEE/RSJ International Conference on Intelligent Robots and
  Systems (IROS)}, pp. 2135--2142, 2017.

\bibitem{Bircher2016}
A.~Bircher, M.~Kamel, K.~Alexis, H.~Oleynikova, and R.~Siegwart, ``Receding
  horizon ``next-best-view" planner for {3D} exploration,'' in \emph{2016 IEEE
  International Conference on Robotics and Automation (ICRA)}, pp. 1462--1468,
  2016.

\bibitem{Hongbiao2021}
H.~Zhu, C.~Cao, Y.~Xia, S.~Scherer, J.~Zhang, and W.~Wang, ``{DSVP}: Dual-stage
  viewpoint planner for rapid exploration by dynamic expansion,'' in \emph{2021
  IEEE/RSJ International Conference on Intelligent Robots and Systems (IROS)},
  pp. 7623--7630, 2021.

\bibitem{Zhong2022}
P.~Zhong, B.~Chen, S.~Lu, X.~Meng, and Y.~Liang, ``Information-driven fast
  marching autonomous exploration with aerial robots,'' \emph{IEEE Robotics and
  Automation Letters}, vol.~7, no.~2, pp. 810--817, 2022.

\bibitem{Chaoqun2020}
C.~Wang, H.~Ma, W.~Chen, L.~Liu, and M.~Q.-H. Meng, ``Efficient autonomous
  exploration with incrementally built topological map in {3-D} environments,''
  \emph{IEEE Transactions on Instrumentation and Measurement}, vol.~69, no.~12,
  pp. 9853--9865, 2020.

\bibitem{Dharmadhikari2020}
M.~Dharmadhikari, T.~Dang, L.~Solanka, J.~Loje, H.~Nguyen, N.~Khedekar, and
  K.~Alexis, ``Motion primitives-based path planning for fast and agile
  exploration using aerial robots,'' in \emph{2020 IEEE International
  Conference on Robotics and Automation (ICRA)}, pp. 179--185, 2020.

\bibitem{Boyu2021}
B.~Zhou, Y.~Zhang, X.~Chen, and S.~Shen, ``{FUEL}: Fast {UAV} exploration using
  incremental frontier structure and hierarchical planning,'' \emph{IEEE
  Robotics and Automation Letters}, vol.~6, no.~2, pp. 779--786, 2021.

\bibitem{Yinghao2024}
Y.~Zhao, L.~Yan, H.~Xie, J.~Dai, and P.~Wei, ``Autonomous exploration method
  for fast unknown environment mapping by using {UAV} equipped with limited fov
  sensor,'' \emph{IEEE Transactions on Industrial Electronics}, vol.~71, no.~5,
  pp. 4933--4943, 2024.

\bibitem{Boyu2023}
B.~Zhou, H.~Xu, and S.~Shen, ``{RACER}: Rapid collaborative exploration with a
  decentralized {Multi-UAV} system,'' \emph{IEEE Transactions on Robotics},
  vol.~39, no.~3, pp. 1816--1835, 2023.

\bibitem{Yichen2024}
Y.~Zhang, X.~Chen, C.~Feng, B.~Zhou, and S.~Shen, ``{FALCON}: Fast autonomous
  aerial exploration using coverage path guidance,'' \emph{IEEE Transactions on
  Robotics}, vol.~41, pp. 1365--1385, 2025.

\bibitem{Song2018}
S.~Song and S.~Jo, ``Surface-based exploration for autonomous {3D} modeling,''
  in \emph{2018 IEEE International Conference on Robotics and Automation
  (ICRA)}, pp. 4319--4326, 2018.

\bibitem{Drew2024}
D.~Hanover, A.~Loquercio, L.~Bauersfeld, A.~Romero, R.~Penicka, Y.~Song,
  G.~Cioffi, E.~Kaufmann, and D.~Scaramuzza, ``Autonomous drone racing: A
  survey,'' \emph{IEEE Transactions on Robotics}, vol.~40, pp. 3044--3067,
  2024.

\bibitem{Cao2025}
Y.~Cao, J.~Lew, J.~Liang, J.~Cheng, and G.~Sartoretti, ``{DARE}: Diffusion
  policy for autonomous robot exploration,'' in \emph{2025 IEEE International
  Conference on Robotics and Automation (ICRA)}, pp. 11\,987--11\,993, 2025.

\bibitem{Cao2023}
Y.~Cao, T.~Hou, Y.~Wang, X.~Yi, and G.~Sartoretti, ``{ARiADNE}: A reinforcement
  learning approach using attention-based deep networks for exploration,'' in
  \emph{2023 IEEE International Conference on Robotics and Automation (ICRA)},
  pp. 10\,219--10\,225, 2023.

\bibitem{chaplotlearning}
D.~S. Chaplot, D.~Gandhi, S.~Gupta, A.~Gupta, and R.~Salakhutdinov, ``Learning
  to explore using active neural {SLAM},'' in \emph{International Conference on
  Learning Representations}.

\bibitem{Tordesillas2022}
J.~Tordesillas, B.~T. Lopez, M.~Everett, and J.~P. How, ``{FASTER}: Fast and
  safe trajectory planner for navigation in unknown environments,'' \emph{IEEE
  Transactions on Robotics}, vol.~38, no.~2, pp. 922--938, 2022.

\bibitem{Selin2019}
M.~Selin, M.~Tiger, D.~Duberg, F.~Heintz, and P.~Jensfelt, ``Efficient
  autonomous exploration planning of large-scale {3-D} environments,''
  \emph{IEEE Robotics and Automation Letters}, vol.~4, no.~2, pp. 1699--1706,
  2019.

\bibitem{Batinovic2022}
A.~Batinovic, A.~Ivanovic, T.~Petrovic, and S.~Bogdan, ``A shadowcasting-based
  next-best-view planner for autonomous {3D} exploration,'' \emph{IEEE Robotics
  and Automation Letters}, vol.~7, no.~2, pp. 2969--2976, 2022.

\bibitem{fsmp}
S.~Zhang, X.~Zhang, Q.~Dong, Z.~Wang, H.~Xi, and J.~Yuan, ``{FSMP}: A
  frontier-sampling-mixed planner for fast autonomous exploration of complex
  and large {3-D} environments,'' \emph{IEEE Transactions on Instrumentation
  and Measurement}, pp. 1--1, 2025.

\bibitem{Zezhou2023}
Z.~Sun, B.~Wu, C.~Xu, and H.~Kong, ``Concave-hull induced graph-gain for fast
  and robust robotic exploration,'' \emph{IEEE Robotics and Automation
  Letters}, vol.~8, no.~9, pp. 5528--5535, 2023.

\bibitem{Zhang2024}
S.~Zhang, R.~Cui, W.~Yan, and Y.~Li, ``Dual-layer path planning with pose slam
  for autonomous exploration in gps-denied environments,'' \emph{IEEE
  Transactions on Industrial Electronics}, vol.~71, no.~5, pp. 4976--4986,
  2024.

\bibitem{Tang2023}
B.~Tang, Y.~Ren, F.~Zhu, R.~He, S.~Liang, F.~Kong, and F.~Zhang, ``Bubble
  explorer: Fast uav exploration in large-scale and cluttered 3d-environments
  using occlusion-free spheres,'' in \emph{2023 IEEE/RSJ International
  Conference on Intelligent Robots and Systems (IROS)}, pp. 1118--1125, 2023.

\bibitem{Shuang2025}
S.~Geng, Z.~Ning, F.~Zhang, and B.~Zhou, ``Epic: A lightweight lidar-based aav
  exploration framework for large-scale scenarios,'' \emph{IEEE Robotics and
  Automation Letters}, vol.~10, no.~5, pp. 5090--5097, 2025.

\bibitem{ericson2025information}
L.~Ericson, J.~Pedro, and P.~Jensfelt, ``Information gain is not all you
  need,'' \emph{arXiv preprint arXiv:2504.01980}, 2025.

\bibitem{li2019deep}
H.~Li, Q.~Zhang, and D.~Zhao, ``Deep reinforcement learning-based automatic
  exploration for navigation in unknown environment,'' \emph{IEEE transactions
  on neural networks and learning systems}, vol.~31, no.~6, pp. 2064--2076,
  2019.

\bibitem{gervet2023navigating}
T.~Gervet, S.~Chintala, D.~Batra, J.~Malik, and D.~S. Chaplot, ``Navigating to
  objects in the real world,'' \emph{Science Robotics}, vol.~8, no.~79, p.
  eadf6991, 2023.

\bibitem{dong2024fast}
Q.~Dong, H.~Xi, S.~Zhang, Q.~Bi, T.~Li, Z.~Wang, and X.~Zhang, ``Fast and
  communication-efficient multi-{UAV} exploration via voronoi partition on
  dynamic topological graph,'' in \emph{2024 IEEE/RSJ International Conference
  on Intelligent Robots and Systems (IROS)}, pp. 14\,063--14\,070, 2024.

\bibitem{Christofides}
N.~Christofides and C.-M. U. P. P. M. S.~R. GROUP., \emph{Worst-Case Analysis
  of a New Heuristic for the Travelling Salesman Problem}, ser. Management
  sciences research report.\hskip 1em plus 0.5em minus 0.4em\relax Management
  Science[s] Research Group, Graduate School of Industrial Administration,
  Carnegie-Mellon University, 1976.

\bibitem{Hou_2025}
J.~Hou, X.~Zhou, N.~Pan, A.~Li, Y.~Guan, C.~Xu, Z.~Gan, and F.~Gao,
  ``Primitive-swarm: An ultra-lightweight and scalable planner for large-scale
  aerial swarms,'' \emph{IEEE Transactions on Robotics}, pp. 1--20, 2025.

\bibitem{Zhepei2022}
Z.~Wang, X.~Zhou, C.~Xu, and F.~Gao, ``Geometrically constrained trajectory
  optimization for multicopters,'' \emph{IEEE Transactions on Robotics},
  vol.~38, no.~5, pp. 3259--3278, 2022.

\bibitem{He_slam}
D.~He, W.~Xu, N.~Chen, F.~Kong, C.~Yuan, and F.~Zhang, ``Point-{LIO}: Robust
  high-bandwidth light detection and ranging inertial odometry,''
  \emph{Advanced Intelligent Systems}, vol.~5, no.~7, p. 2200459, 2023.

\bibitem{Boyu2019}
B.~Zhou, F.~Gao, L.~Wang, C.~Liu, and S.~Shen, ``Robust and efficient quadrotor
  trajectory generation for fast autonomous flight,'' \emph{IEEE Robotics and
  Automation Letters}, vol.~4, no.~4, pp. 3529--3536, 2019.

\end{thebibliography}

\begin{IEEEbiography}[{\includegraphics[width=0.9in,height=1.2in,clip,keepaspectratio]{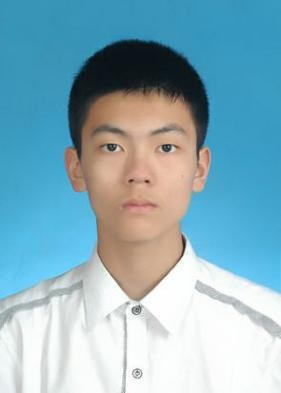}}]
{Qianli Dong} received the B.Eng. degree in  robot engineering from Northeastern University, Shenyang, China, in 2022. He is currently pursuing the Ph.D. degree in control science and engineering with the Institute of Robotics and Automatic Information System, Nankai University, Tianjin, China.

His current research interests include autonomous exploration, visual coverage, and motion planning in unknown environments. 
\end{IEEEbiography}

\begin{IEEEbiography}[{\includegraphics[width=0.9in,height=1.2in,clip,keepaspectratio]{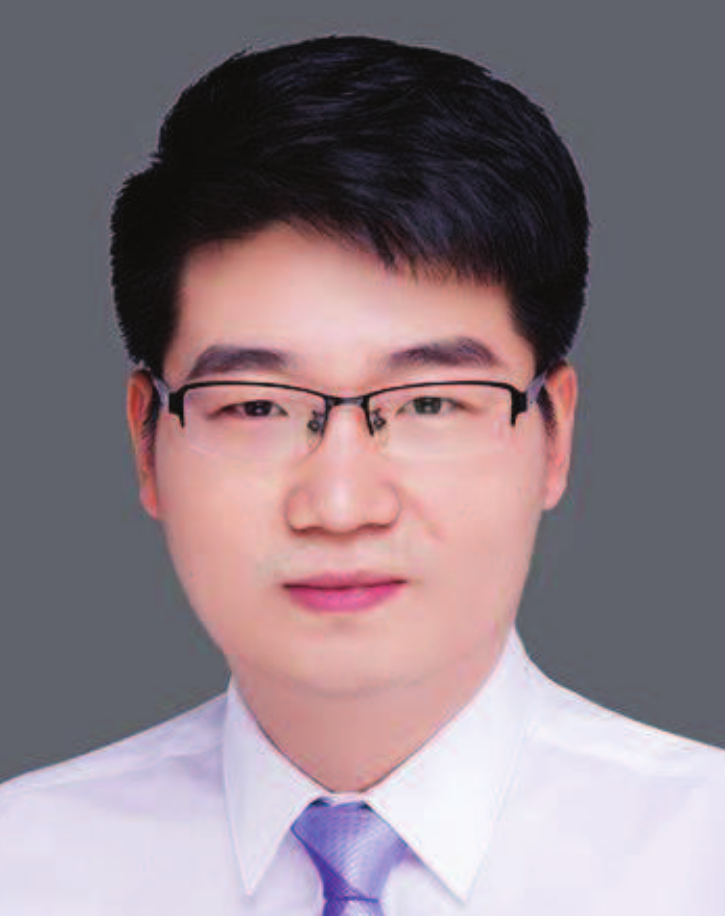}}]
{Xuebo Zhang} (M'12-SM'17) received the B.Eng. degree in automation from Tianjin University, Tianjin, China, in 2006, and the Ph.D. degree in control theory and control engineering from Nankai University, Tianjin, China, in 2011. From 2014 to 2015, he was a Visiting Scholar with the Department of Electrical and Computer Engineering, University of Windsor, Windsor, ON, Canada. He was a Visiting Scholar with the Department of Mechanical and Biomedical Engineering, City University of Hong Kong, Hong Kong, in 2017. He is currently a Professor with the Institute of Robotics and Automatic Information System, Nankai University, and Tianjin Key Laboratory of Intelligent Robotics, Nankai University. His research interests include planning and control of autonomous robotics and mechatronic system with focus on timeoptimal planning and visual servo control; intelligent perception including robot vision, visual servo networks, SLAM; reinforcement learning and game theory.

Dr. Zhang is a Technical Editor of \emph{IEEE/ASME Transactions on Mechatronics} and an Associate Editor of \emph{ASME Journal of Dynamic Systems, Measurement, and Control}.
\end{IEEEbiography}

\begin{IEEEbiography}[{\includegraphics[width=0.9in,height=1.2in,clip,keepaspectratio]{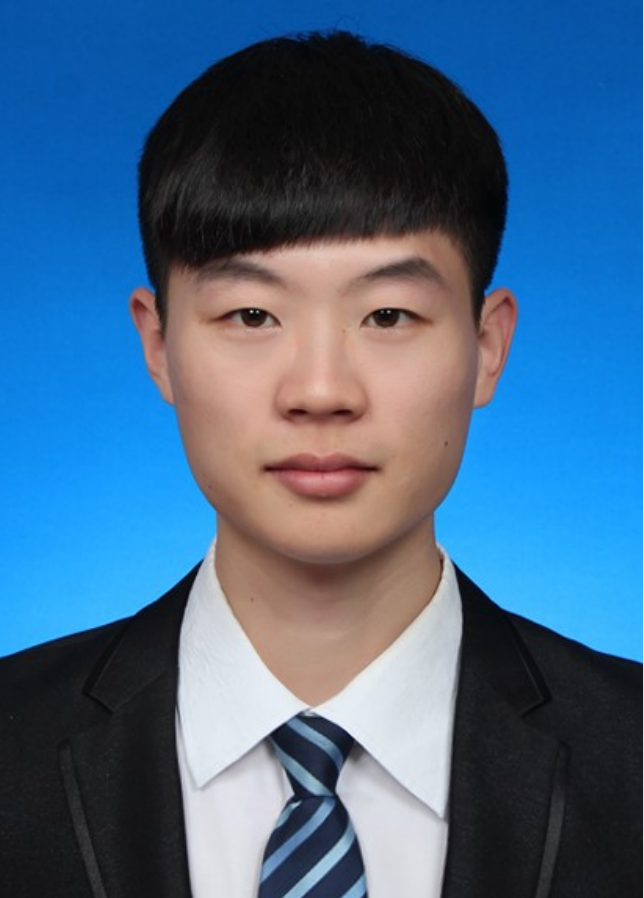}}]
{Shiyong Zhang} received the B.Eng. degree in electrical engineering and automation from Northeast Forestry University, Harbin, China, in 2017, and the Ph.D. degree in control science and engineering from Nankai University, Tianjin, China, in 2022. He is currently a Post-Doctoral Fellow with the Institute of Robotics and Automatic Information System (IRAIS) and also Tianjin Key Laboratory of Intelligent Robotics, Nankai University, China. 

His current research interests include micro aerial vehicles, autonomous exploration, informative path planning, and motion planning in complex environments.
\end{IEEEbiography}

\begin{IEEEbiography}[{\includegraphics[width=0.9in,height=1.2in,clip,keepaspectratio]{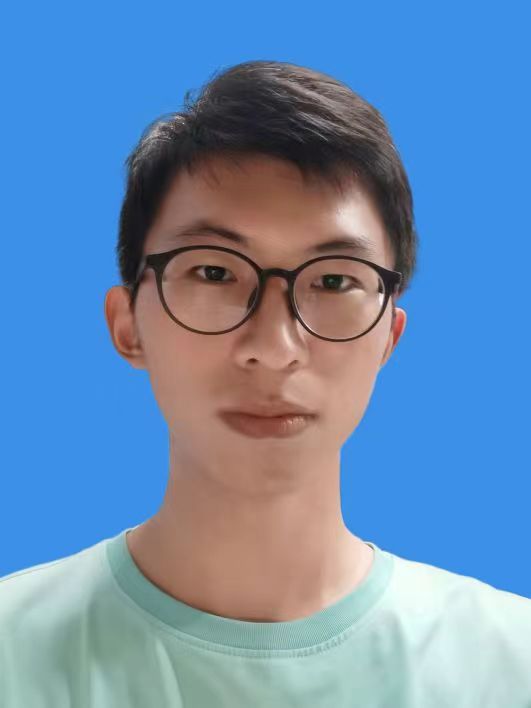}}]
{Ziyu Wang} received the B.Eng. degree in intelligent engineering and technology from Nankai University, Tianjin, China, in 2024. He is currently pursuing the master degree in control science and engineering with the Institute of Robotics and Automatic Information System, Nankai University, Tianjin, China.

His current research interests include unmanned aerial vehicles, motion planning in unknown environments.
\end{IEEEbiography}

\begin{IEEEbiography}[{\includegraphics[width=0.9in,height=1.2in,clip,keepaspectratio]{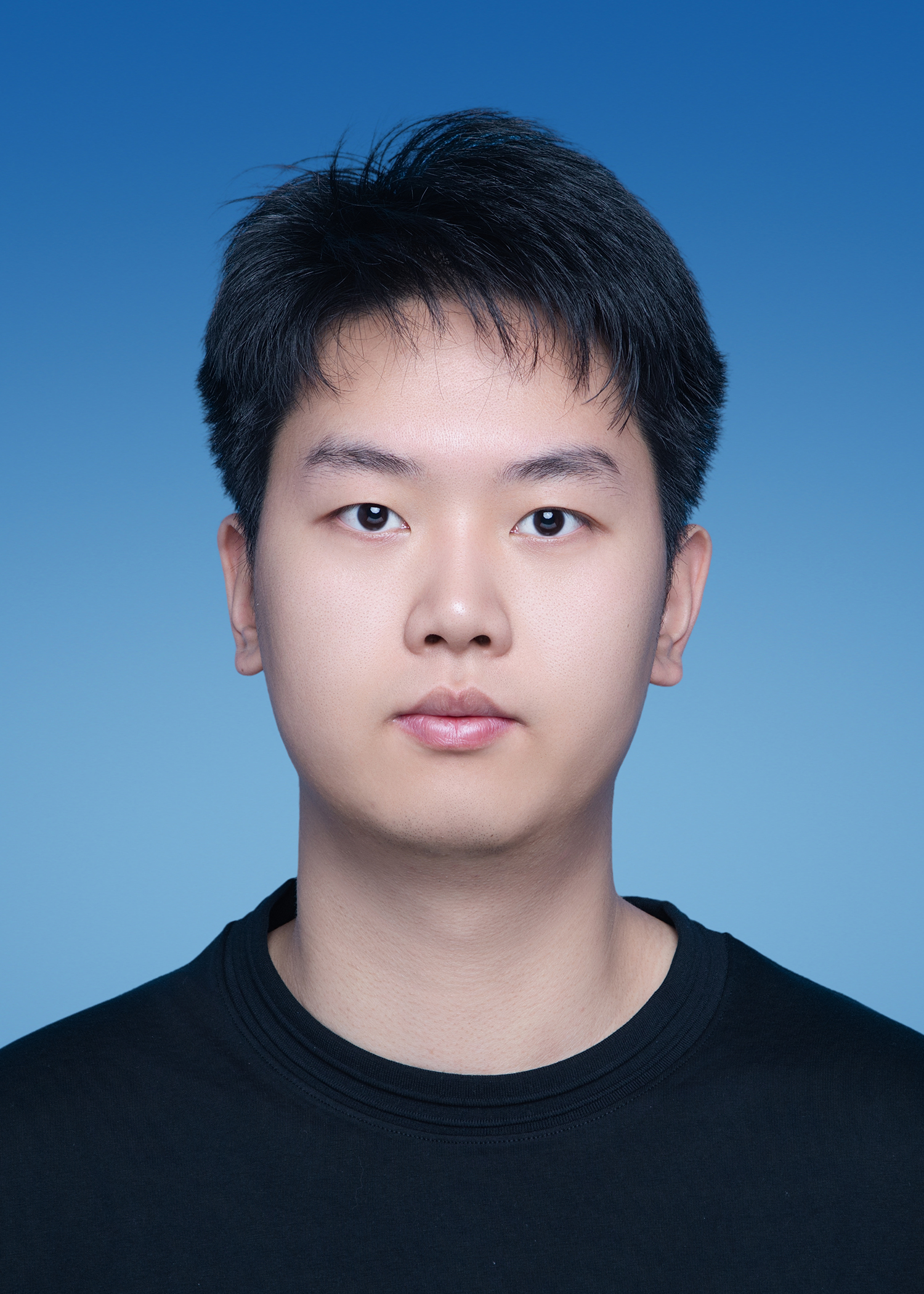}}]
{Zhe Ma} received the B.Eng. degree in intelligent engineering and technology from Nankai University, Tianjin, China, in 2025. He is currently pursuing the master degree in control science and engineering with the Institute of Robotics and Automatic Information System, Nankai University, Tianjin, China. 

His current research focuses on the integration of unmanned aerial vehicles with multimodal large models.
\end{IEEEbiography}

\begin{IEEEbiography}[{\includegraphics[width=0.9in,height=1.2in,clip,keepaspectratio]{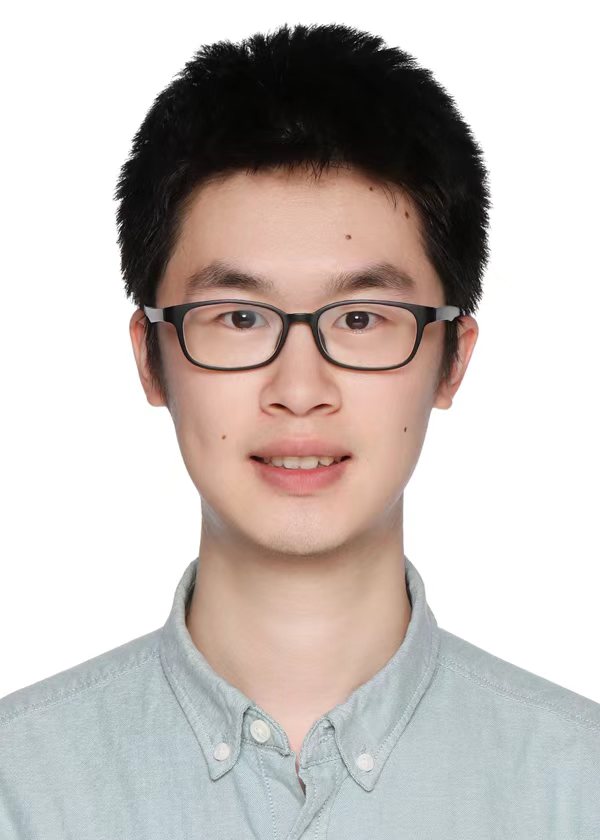}}]
{Haobo Xi} received the B.Eng. degree in artificial intelligence from Dalian University of Technology, Dalian, China, in 2023. He is currently working toward the master's degree in artificial intelligence with the Institute of Robotics and Automatic Information System, Nankai University, Tianjin, China.

His current research interests include unmanned aerial vehicles, LiDAR-based simultaneous localization and mapping.
\end{IEEEbiography}

\end{document}